\documentclass[journal]{IEEEtran}
\IEEEoverridecommandlockouts
\usepackage{cite}
\usepackage{amsmath,amssymb,amsfonts}
\usepackage{algorithmic}
\usepackage{graphicx}
\usepackage{textcomp}
\usepackage{xcolor}

\usepackage{multirow}
\usepackage{booktabs}
\usepackage{multicol}
\usepackage{adjustbox}

\usepackage{pifont} 
\newcommand{\cmark}{\ding{51}} 
\newcommand{\xmark}{\ding{55}} 
\usepackage{lipsum} 

\usepackage[ruled,linesnumbered]{algorithm2e}

\def\BibTeX{{\rm B\kern-.05em{\sc i\kern-.025em b}\kern-.08em
    T\kern-.1667em\lower.7ex\hbox{E}\kern-.125emX}}

\newcommand{\biaoti}{\fontsize{23.7pt}{\baselineskip}\selectfont}

\begin{document}

\title{\biaoti Heterogeneity-aware Personalized Federated Learning via Adaptive Dual-Agent Reinforcement Learning}

\author{
    Xi Chen,~\IEEEmembership{Student Member,~IEEE,}
    Qin~Li,~\IEEEmembership{Member,~IEEE,}
    Haibin Cai, 
    Ting~Wang,~\IEEEmembership{Senior Member,~IEEE}
\IEEEcompsocitemizethanks{
    \IEEEcompsocthanksitem Xi Chen, Qin Li, Haibin Cai, and Ting Wang are with the MoE Engineering Research Center of Software/Hardware Co-design Technology and Application; Shanghai Key Laboratory of Trustworthy Computing; and East China Normal University, Shanghai, 200062, China. (Emal: 71255902127@stu.ecnu.edu.cn, \{qli, hbcai, twang\}@sei.ecnu.edu.cn).
    The corresponding author is Ting Wang.
}


}

\maketitle

\begin{abstract}

Federated Learning (FL) empowers multiple clients to collaboratively train machine learning models without sharing local data, making it highly applicable in heterogeneous Internet of Things (IoT) environments. However, intrinsic heterogeneity in clients' model architectures and computing capabilities often results in model accuracy loss and the intractable straggler problem, which significantly impairs training effectiveness.
To tackle these challenges, this paper proposes a novel Heterogeneity-aware Personalized Federated Learning method, named HAPFL, via multi-level Reinforcement Learning (RL) mechanisms. HAPFL optimizes the training process by incorporating three strategic components: 1) An RL-based heterogeneous model allocation mechanism. The parameter server employs a Proximal Policy Optimization (PPO)-based RL agent to adaptively allocate appropriately sized, differentiated models to clients based on their performance, effectively mitigating performance disparities. 2) An RL-based training intensity adjustment scheme. The parameter server leverages another PPO-based RL agent to dynamically fine-tune the training intensity for each client to further enhance training efficiency and reduce straggling latency. 3) A knowledge distillation-based mutual learning mechanism. Each client deploys both a heterogeneous local model and a homogeneous lightweight model named LiteModel, where these models undergo mutual learning through knowledge distillation.
This uniform LiteModel plays a pivotal role in aggregating and sharing global knowledge, significantly enhancing the effectiveness of personalized local training.
Experimental results across multiple benchmark datasets demonstrate that HAPFL not only achieves high accuracy but also substantially reduces the overall training time by 20.9\%-40.4\% and decreases straggling latency by 19.0\%-48.0\% compared to existing solutions.


\end{abstract}


\section{Introduction}
With the continuous advancement of Internet of Things (IoT) technology \cite{al2015internet}, the deployment of IoT devices is experiencing unprecedented growth. As of 2021, there are already over 10 billion active IoT devices worldwide, and this number is expected to exceed 25.4 billion by 2030. By 2025, an estimated 152,200 IoT devices will be connecting to the Internet every minute, contributing to a projected increase in data generation to 73.1 Zettabytes by that year—a dramatic 422\% rise from the 17.3 Zettabytes recorded in 2019.
Traditionally, such vast amounts of data have been managed through centralized learning (CL) methods \cite{lecun1998gradient} \cite{krizhevsky2012imagenet}, which aggregate data to train or fine-tune models. Although CL is effective for developing high-precision models and straightforward to implement, it relies heavily on centralized data collection. This dependency raises substantial privacy and security concerns, particularly as IoT devices often hold sensitive information, such as medical records containing patients' private health data or personal sensitive banking details. Data owners' reluctance to share such data compromises the feasibility of CL for practical applications. Moreover, the CL model, necessitating extensive data collection, incurs significant communication costs, which make it impractical for resource-constrained devices operating in  bandwidth-limited edge wireless networks, thereby restricting the scalability and applicability of the CL approach in the rapidly evolving IoT environment.

Federated Learning (FL) embodies a collaborative learning paradigm, specifically designed to address privacy concerns. 
This approach utilizes a distributed learning algorithm that allows clients to independently process local datasets without the necessity to exchange raw data \cite{10264807}. 
In FL system, the server distributes a global model to each participating client, who then utilizes their unique local dataset to train a local model. 
Upon completion of training, clients transmit their model parameters to the server, which aggregates these contributions to refine the global model. This strategy ensures that the global model can effectively leverage the collective data from all participating clients without compromising individual data privacy, thus making FL highly practical and broadly applicable across various data-sensitive sectors.


Despite its vast potential, FL encounters significant implementation challenges that hinder its widespread adoption. In addition to managing data privacy, one primary issue in FL arises from the computational and communication efficiency constraints that occur within networks of heterogeneous devices \cite{bonawitz2019towards} \cite{yang2022over} \cite{yang2023decentralized}. In synchronous FL systems, performance disparity—defined here as any factors that impact client training in a given round and contribute to training latency—can include specific elements such as computational resources, dataset size, and model size. This disparity often leads to “straggler problems,” where high-performance devices must wait for slower devices to complete their tasks before proceeding to aggregation. This delay can severely impact resource utilization and model efficiency, underscoring the need for methods that can dynamically adapt to the varied capacities of different clients.


In synchronous FL, the disparity in performance among heterogeneous devices often causes ``straggler problems", where high-performance devices are delayed as they must wait for lower-performance devices to complete training and data upload before aggregation can proceed \cite{dean2013tail}\cite{dean2008mapreduce}. One common strategy to mitigate this issue and reduce straggling latency involves allocating different training intensities based on client performance, where ``training intensity" in this context refers to varying the number of training epochs assigned to each client during federated training rounds.
However, while this approach can enhance the efficiency of  high-performance devices, it also risks exacerbating disparities. Specifically, allocating too many epochs to high-performance devices and too few to lower-performance ones can lead to an imbalance, which might not only decrease the accuracy of the aggregated model due to under-representation of data from less capable devices but also result in suboptimal utilization of all client data across the network.

While Semi-FL \cite{stripelis2022semi} and Asyn-FL \cite{xu2024enhancing} approaches have been proposed to address training latency in federated learning, each comes with notable limitations. Asyn-FL improves scalability and efficiency by allowing clients to update models asynchronously, yet it often struggles with stability in model convergence, especially in environments with numerous nodes or significant computational delays. Similarly, Semi-FL seeks a balance between global consistency and performance efficiency by introducing a delay window for client updates. However, its effectiveness depends heavily on carefully tuning this window, which complicates deployment in large-scale systems.

To address the challenges outlined above, this paper introduces a novel adaptive Heterogeneity-aware Personalized Federated Learning method, named HAPFL. HAPFL leverages two deep reinforcement learning (DRL) agents to dynamically allocate appropriately sized heterogeneous models and tailored training intensities to clients based on their performance capabilities, respectively. Furthermore, to tackle the obstacles posed by model heterogeneity during global model aggregation, each client is additionally equipped with a homogeneous lightweight model, named LiteModel. This setup allows each client to train its personalized local model under intensities specified by the reinforcement learning (RL) agent, while the LiteModel engages in mutual learning with the local model through knowledge distillation \cite{luo2024federated}. These well-learned homogeneous LiteModels are then used to perform global model aggregation. This dual-model approach not only facilitates efficient and effective global model aggregation but also enhances the local model with insights gained from the LiteModel, which accumulates and transfers knowledge across all clients. 
Moreover, to mitigate the adverse impacts of low-contributing clients on the accuracy and stability of the global model, this paper proposes a novel model aggregation method using information entropy and accuracy weighting.
In summary, this paper makes the following three key contributions:
\begin{enumerate}
    \item We propose a novel heterogeneity-aware personalized federated learning framework, named HAPFL, that leverages two functional RL agents. These two agents are respectively designed to adaptively determine appropriately sized heterogeneous models for each client and dynamically adjust each client's training intensities based on their computing capabilities and performance, aiming to effectively mitigate the serious straggler problem and reduce the straggling latency. 
    \item We introduce a lightweight homogeneous model called LiteModel, which is deployed on each client. This LiteModel and the corresponding local model on each client engage in continuous knowledge transfer through knowledge distillation-based mutual learning. The LiteModel serves as a universally consistent model designed to aggregate and distribute global knowledge, thereby facilitating local training processes and effectively tackling the challenges associated with heterogeneous models.
    \item We develop a prototype of HAPFL and conduct extensive simulations to evaluate its performance on three well-known datasets: MNIST, CIFAR-10 and ImageNet-10. Experimental results demonstrate that our HAPFL approach significantly outperforms baseline methods, improving model accuracy by up to 7.3\%, reducing overall training time by 20.9\% to 40.4\%, and decreasing straggling latency differences by 19.0\% to 48.0\%.
\end{enumerate}

The remainder of this paper is organized as follows. Section \ref{related_work} briefly reviews the related works. Section \ref{system_model} presents the system modeling and problem formulation. Section \ref{Approach_Design} elaborates on the design of our HAPFL approach. Section \ref{evaluation} provides experimental results. Section \ref{conclusion} concludes this paper.

\section{Related Work}
\label{related_work}
This section briefly reviews some related work in FL and discusses the limitations of these methods.

The most widely recognized algorithm in FL is the Federated Average (FedAvg) \cite{cho2023communication} proposed by Google. However, FedAvg is built upon assumptions of independent and identically distributed (IID) data, uniform model architectures, and reliable network connections. In practice, FedAvg struggles with non-IID data and client performance heterogeneity, which is common in real-world scenarios \cite{liu2022towards}.
To address the issue of heterogeneous client performance, various methods have been proposed. Lei Yang et al. \cite{yang2023personalized} utilized a clustering approach that groups clients based on performance, allocating the same model to each cluster and aggregating the different models on the server side.
Ruixuan Liu et al. \cite{liu2022no} introduced a hierarchical heterogeneous aggregation framework that assigns different-sized models to clients based on their computing capabilities, with a hierarchical aggregation of global models. 
Jun Xia et al. \cite{xia2022pervasivefl} adopted lightweight modellets and local models of varying sizes, training them through Deep Mutual Learning and aggregating all modellets on the server side for knowledge sharing. 
Yae Jee Cho et al. \cite{cho2023communication} tackled model heterogeneity by passing soft labels of local models, clustering soft labels on the server side, and training local models through knowledge distillation. 
Jianyu Wang et al. \cite{wang2020tackling} proposed a method to address the objective inconsistency problem in heterogeneous federated optimization. By normalizing the model updates from clients, FedNova ensures comparability of updates from clients with diverse data distributions and computational capacities, enhancing convergence stability in federated learning settings with significant client heterogeneity.
While these methods improve client performance utilization, they rely on pre-allocated fixed models and training intensities. In dynamic client groups with significant performance disparities, these approaches may still result in high straggler latency.

Personalized Federated Learning (PFL) aims to address client heterogeneity by tailoring models to individual clients. For example, Smith et al. \cite{smith2017federated} proposed Federated Multi-task Learning, where different tasks were learned across clients. Muhammad et al. \cite{muhammad2020fedfast} introduced Fedfast, a framework designed to accelerate federated recommender system training by allocating different training intensities to clients based on their capabilities, which reduces latency and improves efficiency. Li et al. \cite{li2021ditto} developed Ditto, which used personalized aggregation schemes to improve fairness and robustness in federated settings. Similarly, Fallah et al. \cite{fallah2020personalized} leveraged meta-learning techniques to personalize the model for each client. These efforts aim to account for individual differences across clients by customizing models based on local data and local conditions. However, despite the success of these methods in accommodating client heterogeneity, they do not directly address the issue of client latency and the associated straggler problem. In dynamic environments where clients experience varying latencies and computational capabilities, these personalized approaches may still suffer from inefficiencies, as clients with slower performance can cause delays in the overall training process, leading to high straggler latency.

Recent advancements in PFL have proposed methods that go beyond basic personalization. For example, Deng et al. \cite{deng2024fedasa} introduced FedASA, which adapts model aggregation to account for heterogeneous mobile edge computing environments, addressing some aspects of latency and model convergence. Lee et al. \cite{lee2024fedl2p} proposed FedL2P, which personalizes federated learning by utilizing model regularization techniques to handle data heterogeneity. Yang et al. \cite{yang2024fedas} further extended these ideas by developing FedAS, a method that bridges inconsistencies in personalized federated learning. These methods, while advancing PFL, still do not fully tackle the latency issues faced by clients with significantly varying performance capabilities. In these scenarios, where high-performance clients finish training faster than low-performance ones, the straggler problem persists.

Moreover, in FL, clients' training intensity is mostly uniform. However, in synchronous FL, high-performance clients completing training more quickly than lower-performance ones exacerbate the ``straggler problem," where faster clients must wait for slower ones before global aggregation can proceed.
Although asynchronous FL \cite{liu2024aedfl}  \cite{chen2020asynchronous}\cite{lian2018asynchronous} presents advantages in addressing the straggler problem, it also poses risks of model quality degradation and server crashes. 
Several approaches have been proposed to address this issue. Tianming Zang et al. \cite{zang2023general} implemented a clustering method based on upload or communication time \cite{xu2023enhancing}, ensuring that clients with similar time metrics are grouped together. In each training round, clients within the same group are selected for training, thereby reducing waiting time and mitigating the straggler problem. Yangguang Cui et al. \cite{cui2022optimizing} introduced a utility-driven and heterogeneity-aware heuristic user selection strategy. By considering client performance and heterogeneity, this approach optimizes client selection to minimize straggling latency. Peichun Li et al. \cite{li2023anycostfl} proposed on-demand learning, which adjusts the local model structure, gradient compression strategy, and computational frequency according to personalized latency and energy constraints. 

Although these methods effectively reduce straggling latency, they may face challenges in adapting to complex and dynamic environments. Further research is needed to develop robust and adaptable solutions for addressing the straggler problem in FL effectively. In recent years, DRL
has emerged as a powerful tool for solving complex sequence decision optimization problems \cite{gupta2021embodied}. Given that FL can be conceptualized as a Markov decision process (MDP), DRL presents a promising approach for optimizing FL. Building on the success of DRL in FL optimization, several studies have applied DRL to address FL resource allocation challenges. To tackle the straggler problem in FL, Manying Zeng et al. \cite{zeng2022heterogeneous} utilized a DRL model to dynamically adjust local training intensity. By adaptively modifying the training intensity based on local conditions, this approach effectively reduces straggling latency, enhancing overall FL performance. Similarly, Yi Jie Wong et al. \cite{wong2023fedddrl} employed a DRL model for client selection and allocation of training intensity. By leveraging DRL, clients are intelligently selected and assigned varying levels of training intensity, thereby optimizing FL resource allocation and further mitigating the straggler problem. 
While these methods have shown effectiveness in reducing straggling latency, they face challenges when dealing with client groups exhibiting large performance disparities. Extreme resource allocation in such scenarios could potentially lead to performance degradation of the global model. When performance differences among clients are significant, allocating extensive training intensity to high-performance clients while providing minimal training intensity to lower-performance clients can create extreme imbalances in training. This imbalance may lead to the global model’s performance degradation, as the limited training experienced by lower-performance clients can result in insufficient learning and poor model convergence. Thus, further research is essential to develop robust and adaptive DRL-based approaches that can effectively handle varying performance levels among client groups in FL settings. In this context, in this paper, we propose a multi-level DRL-based personalized FL framework that incorporates two DRL agents to adaptively determine the client model sizes and training intensities for each client, respectively, and introduce a homogeneous lightweight LiteModel for global model aggregation enhanced by knowledge distillation-based mutual learning to address the model heterogeneity issues. Table \ref{tab:method_comparison} lists some comparisons between previous works and our work.

\begin{table}[h!]
\centering
\caption{Comparison of Methods on Statistical Heterogeneity, Resource Constraints, and Straggler Problem}
\label{tab:method_comparison}
\setlength{\tabcolsep}{4pt} 
\resizebox{\columnwidth}{!}{ 
\begin{tabular}{lccc}
\toprule
\textbf{Method}    & \textbf{Statistical Heterogeneity} & \textbf{Resource Constraints} & \textbf{Straggler Problem} \\ 
\midrule
\textbf{FedAvg}    & \xmark                            & \xmark                       & \xmark                     \\
\textbf{FedProx}   & \cmark                            & Partial \cmark                       & \xmark                     \\
\textbf{pFedMe}    & \cmark                            & Partial \cmark                       & \xmark                     \\
\textbf{Ditto}     & \cmark                            & \xmark                       & \xmark                     \\
\textbf{FedASA}    & \cmark                            & \cmark                       & \xmark                     \\
\textbf{FedNova}   & \cmark                            & Partial \cmark                       & \cmark                     \\
\textbf{FedDdrl}   & \cmark                            & Partial \cmark                       & \cmark                     \\
\textbf{HAPFL}     & \cmark                            & \cmark                       & \cmark                     \\ 
\bottomrule
\end{tabular}
}
\end{table}

\section{System Model and Problem Formulation}
\label{system_model}
This section describes our HAPFL system model, mainly consisting of five phases, and the problem formulation that will be addressed in this paper.

\subsection{System Model}

Consider a FL system comprising $ K $ devices, each capable of participating in the learning process. In each communication round  $ r $, a subset of $ k $ clients is randomly selected from the total pool of $ K $ clients. Specifically, each round includes the following five phases:

\textbf{Performance Assessment Training}: 
Each client performs training with the LiteModel $w^{lite}_{r,i}$ to evaluate their current computational capabilities.
The resulting data, which reflects each client's performance capability, is then sent back to the server.
This step is crucial as it provides the server with each client's real-time processing power and readiness, enabling the server to make better decisions about allocations of model sizes and training intensities in subsequent phases.
\begin{equation}
    P_{r,i}=f_{eval}(w^{lite}_{r,i};\mathbb{D}_i),
\end{equation}
where $ P_{r,i} $ denotes the performance metric of client $ i $ at round $ r $, $ f_{eval} $ is the assessment function, and $ \mathbb{D}_i $ represents the local dataset of client $ i $.

\textbf{Adaptive Training Adjustment}: 
The server employs two distinct Proximal Policy Optimization (PPO)-based RL agents. Utilizing the performance assessment information $ P_{r} $ received from clients, these two agents generate two separate policies: $\pi_s$ and $\pi_E$. The first policy $\pi_s$ determines the optimal size $ s^{size}_{r} $ of each client's local model, while the second policy $\pi_E$ adjusts the training intensities, specifically the number of training epochs $ E_{r} $. These tailored configurations are then communicated back to the clients for implementation.
\begin{equation}
    \pi_s:P_{r} \rightarrow s^{size}_{r};\quad
    \pi_E:P_{r} \rightarrow E_{r}.
\end{equation}
These training adjustments are specifically tailored based on the latest insights derived from RL agents, using the current performance metrics of clients. This ensures that all clients perform local training with models and training intensities optimally suited to their distinct capacities.






\textbf{Model Distribution}: 
After making adaptive training adjustments with two RL agents in round $r$, the server dispatches the most recent global LiteModel $ w^{lite}_{r} $, along with the corresponding aggregated differentiated local models $ w^{local}_{r} $, to the participating clients based on the adjustment information. This ensures that all clients start their local training processes with the most updated models, maintaining consistency across the learning network.

\textbf{Local Training and Mutual Learning}: 
Clients initialize or update their LiteModels $w^{lite}_{r,i}$ and local models $w^{local}_{r,i}$, as well as the assigned number of training epochs $E_{r,i}$, according to the decisions issued by the PPO agents. Subsequently, both the LiteModel $w^{lite}_{r,i}$ and the local model $w^{local}_{r,i}$ engage in local training using local datasets $ \mathbb{D}_i$ while simultaneously participating in mutual learning through knowledge distillation. After the training phase concludes, clients upload their newly updated local model and LiteModel back to the server. The model is updated using the stochastic gradient descent algorithm, as follows:
\begin{equation}
    w_{r+1,i} = w_{r,i} - \eta \nabla L_i(w_{r,i}, \mathbb{D}_i),
\end{equation}
where $ \eta $ is the learning rate and $ L_i $ is the loss function.

\textbf{Global Model Aggregation}: 
The server aggregates the LiteModels $ \{w^{lite}_{r+1,i}\}^k_{i=1} $ received from all clients to update the global LiteModel:
\begin{equation}
    w^{lite}_{r+1} = \frac{1}{k} \sum^k_{i=1} w^{lite}_{r+1,i}.
\end{equation}
Additionally, it organizes the variously sized local models into several groups based on their sizes and aggregates the same-sized models within each group separately:
\begin{equation}
    w^{(size)}_{r+1} = \frac{1}{|{\mathcal{G}^{(size)}_{r+1}}|} \sum_{i \in \mathcal{G}^{(size)}_{r+1}} w^{local}_{r+1,i},
\end{equation}
where $ \mathcal{G}^{(size)}_{r+1} $ is the group of clients with the same model size $ size $.
This dual model aggregation process prepares them for the next round of each client's model initialization, ensuring that updates are efficiently tailored to each client's diverse capabilities and configurations.


This improved FL system facilitates collaborative training of heterogeneity-aware personalized models across distributed heterogeneous devices while preserving data privacy. The adaptive training strategies ensure that each client contributes effectively to the collective learning process while realizing each client model's personalization. Notably, the incorporation of LiteModels facilitates a dual-learning mechanism that boosts not only personalized local model performance through knowledge distillation but also enhances global model accuracy through knowledge transfer and systematic aggregation.


\begin{figure*} 
  \centering
  \includegraphics[width=0.8\linewidth]{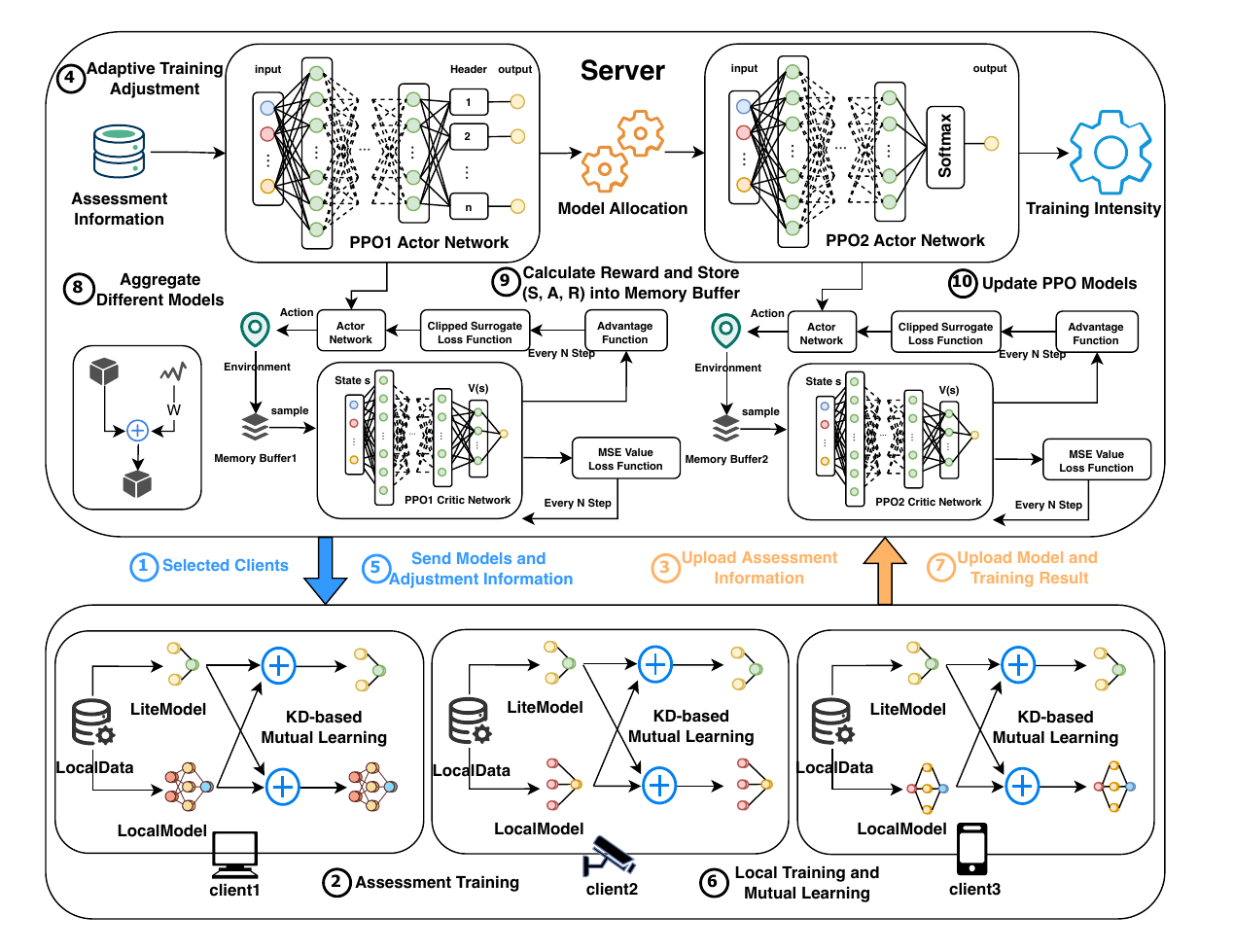} 
  \caption{Overview of HAPFL architecture}
  \label{fig:systemModel}
  \vspace{-0.3cm}
\end{figure*}

\subsection{Problem Formulation}
In a typical FL system comprising a server and $ K $ clients, each client $ i \in K $ maintains a local dataset $ \mathbb{D}_i $ of size $ D_i $. The complete FL training process involves several key phases: model distribution, local training, and model aggregation. The total time required for these processes in each communication round $r$ can be approximated as:
\begin{equation}
    T_{r,i} =T^b_{r,i}  + T^c_{r,i} + T^u_{r,i} + T^a_{r,i},
\end{equation}
where $ T^b $ represents the time needed for broadcasting the model to the clients, $ T^c $ means the computation time at the client, $ T^u $ indicates the time spent on model uploading, and $ T^a $ denotes the model aggregation time at the server.

However, the heterogeneity in client capabilities and dataset sizes often results in significant variations in local computation times among clients, leading to significant straggling latency in overall system performance. These straggling latencies are predominantly due to the local computation time, which often exceeds the combined time for model distribution, uploading, and aggregation.


Local computation comprises two components: performance assessment training and local training. Performance assessment training involves one epoch of local training with the client's LiteModel to evaluate the client's computational capacity, with training time denoted as $ T^d $. Local training encompasses one round of training using both the LiteModel and local model on the client side, with training time denoted as $ T^l $. The total time cost for client $i$ in round $r$ is given by:
\begin{equation}
    T^c_{r,i}=T^d_{r,i}+T^l_{r,i}.
\end{equation}

In FL, the server synchronizes model aggregation after all clients complete their training.
Straggling latencies are exacerbated when faster clients must wait for the slower ones.
Defining $\mathrm{S}_r$ as the set of participating clients in the $r$-th training round, with $j$ and $j'$ denoting the slowest and fastest clients, respectively, the straggling latency in the $r$-th round can be expressed as:
\begin{equation}
\label{training_latency}
    \Delta T^c_{r} = T^c_{r,j}-T^c_{r,j'},j \neq j',j\in \mathrm{S}_r,j'\in \mathrm{S}_r,
\end{equation}
\begin{equation}
    T^c_{r,i} = \tau_{r,i} \cdot \hat{T}^e_{r,i},
\end{equation}
\begin{equation}
    \hat{T}^e{r,i} = f(D_i, w^{(size)}_{r,i}, \psi_i),
\end{equation}
where $\hat{T}^e_{r,i}$ is the average time required for a single epoch of training on client $i$. This time depends on the client’s hardware capabilities, dataset size $D_i$, and the size of the allocated model $w^{(size)}{r,i}$. $f(\cdot)$ is a function capturing the client’s computational performance, and $\psi_i$ represents the hardware and resource constraints of client $i$.

Such straggling latency primarily results from differences in client hardware performance and dataset sizes. 
To minimize these latencies, it is effective to allocate models of varying sizes and assign different training intensities based on client performance, which can significantly reduce discrepancies in training times.
Let $\tau_{r,i}$ represent the assigned training intensity for client $i$ in round $r$, with the objective to minimize $\Delta T^c_r$:
\begin{equation}
    \mathop{min}\Delta T^c_r=min(T^c_{r,j}-T^c_{r,j'}),
\end{equation}

Subject to
\begin{equation}
\label{eq:intensityConstraint1}
    \sum_{i \in \mathrm{S}_r}\tau_{r,i} = \tilde{\tau}_r
\end{equation}
\begin{equation}
\label{eq:intensityConstraint2}
    \tau_{r,i} > 0
\end{equation}
\begin{equation}
\label{eq:modelSiseConstraint}
    U(w^{(size)_{r,i}}) \geq \varepsilon
\end{equation}
Equation (\ref{eq:intensityConstraint1}) ensures that the total training intensity allocated to each client in round $ r $ is equivalent to a predefined total $\tilde{\tau}_r$. Equation (\ref{eq:intensityConstraint2}) guarantees that every participating client completes at least one local training iteration. 
Equation (\ref{eq:modelSiseConstraint}) indicates that the parameter count of the allocatable model exceeds the minimum threshold required for the current task, ensuring that the model can effectively fulfill the training task. Here, $U(\cdot)$ represents the function for computing the model parameter count.

\section{HAPFL Approach Design}
\label{Approach_Design}
In this section, we will elaborate on the design of the HAPFL's architecture and detail the specific algorithms employed within this framework. The detailed process of the HAPFL algorithm is shown in Algorithm \ref{HAPFL Algorithm}.

\subsection{Overview of HAPFL Architecture}
The overall architecture of the HAPFL method is illustrated in Figure \ref{fig:systemModel}. In each round of federated training, clients first submit requests to participate in a round of federated training. After receiving these requests, the server selects appropriate clients. 
These selected clients conduct the performance assessment training based on the LiteModel and send the resulting data back to the server. 
Utilizing this assessment information, the server then performs adaptive training adjustments through an RL-based heterogeneous model allocation mechanism and an RL-based training intensity adjustment scheme. These adjustments are designed to optimally balance each client's unique capabilities and overall training efficiency.
The server then sends the LiteModel, heterogeneous local models, and the training adjustment information back to the participating clients. After receiving local models with assigned training intensities, clients undertake local training using a knowledge distillation-based mutual learning mechanism to enhance learning efficiency. Upon completing their training, clients upload the training information to the server, including model parameters, the accuracy of both the local model and LiteModel, the associated training time costs, and the information entropy of the client dataset. 
The server separately aggregates the collected LiteModels and heterogeneous models using a weighted aggregation method that considers both information entropy and model accuracy. Simultaneously, the server calculates RL rewards based on the collected training information and stores the $(states, actions, rewards)$ tuples in a memory buffer. Once the memory buffer is full, it triggers the update of two RL models, continuously refining the training process. This cycle ensures that the HAPFL system dynamically adapts to the evolving needs and capabilities of its diverse clients.

\begin{algorithm}
\label{HAPFL Algorithm}
  \SetKwData{Left}{left}\SetKwData{This}{this}\SetKwData{Up}{up}
  \SetKwFunction{Union}{Union}\SetKwFunction{FindCompress}{FindCompress}
  \SetKwInOut{Input}{input}\SetKwInOut{Output}{output}
  \BlankLine
  \emph{Initialize the FL settings;}\ \\
  \emph{Initialize parameters $\theta$ and $\phi$ in actor network $\pi_{\theta}(\mathcal{A}_r|\mathcal{S}_r)$ and critic network $V_{\phi}(\mathcal{S}_r)$;}\ \\
  \emph{Initialize two experience replay buffers $B_1,B_2$;}\ \\
  \For{episode $e$ in $1, 2, \ldots , E$}{
    \emph{Reset the FL environment;}\ \\
    \For{round $r$ in $1, 2, \ldots , R$}{
        \emph{Randomly select $k$ clients from all $K$ clients;}\ \\
        \For{client $i$ in $1, 2, \ldots , k$}{
            \emph{Perform assessment training;}\ \\
            \emph{Send the $T^d_{r,i}$ to the server;}\ \\
        }
        \emph{Calculate the $\mathcal{S}_r$ and $\mathcal{A}_r$ of two DRL models;}\ \\
        \emph{The server sends the corresponding sized model and training intensities based on the training adjustment results to clients;}\ \\
        \For{client $i$ in $ 1, 2, \ldots, k $}{
            \emph{Set clock to calculate $T^l_{r,i}$;}\ \\
            \emph{Perform local training following Eq. \ref{eq:localModelLossFunction}-\ref{eq:gradientUpdate};}\ \\
            \emph{Calculate $acc_{r,i}$;}\ \\
            \emph{Send $\theta_{r,i}$, $T^l_{r,i}$, $acc_{r,i}$ and $H_{r,i}$ to the server;}\ \\
        }
        \emph{Calculate $W_r$ according to $H_r$ and $acc_r$;}\ \\
        \emph{Update the global model according to $ \theta^{global}_{r+1} \leftarrow \theta^{global}_r + \sum^{k}_{j = 0}W_{r,i} \cdot \theta_{r,i} $;}\ \\
        \emph{Calculate $\mathcal{R}^1_r,\mathcal{R}^2_r$ according to $T^l_r$;}\ \\
        \emph{Store tuple $ (\mathcal{S}^1_r, \mathcal{A}^1_r, \mathcal{R}^1_r) $ into $B_1$;}\ \\
        \emph{Store tuple $ (\mathcal{S}^2_r, \mathcal{A}^2_r, \mathcal{R}^2_r) $ into $B_2$;}\ \\
        \If{buffer $B_1$ is full}{
            \emph{Update PPO1;}\ \\
        }
        \If{buffer $B_2$ is full}{
            \emph{Update PPO2;}\ \\
        }
    }
  }
  \caption{HAPFL Algorithm}\label{algo_disjdecomp}
\end{algorithm}\DecMargin{1em}

\subsection{Performance Assessment Training}

In a real dynamic FL environment, client participation and its available computational resources can vary significantly from round to round. To effectively manage this variability, performance assessment training on each client is employed to provide the server-side RL agents with current computational capabilities of all involved clients.
During the assessment training, clients conduct a brief training with their LiteModel for one epoch, facilitating rapid assessment without significant computational overhead while providing valuable insights into client performance. 
Upone completing this initial training, clients promptly upload the resulting assessment training time cost $T^d_{r,i}$ to the server. This information is crucial for server's RL agents to make informed decisions about model sizes and training intensities for each client in subsequent training rounds, ensuring that the system continuously adapts to the changing dynamics of the client pool and their capabilities.


\subsection{Dual-Agent RL for FL Optimization}
Allocating uniform training intensity to all clients with significant performance differences may suffer from the straggler problem, resulting in high straggling latency.
However, relying solely on adjusting training intensities may lead to imbalanced and inadequate training for some clients, which often results in under-representation of data from lower-performance clients, thereby decreasing the overall accuracy of the model.
To address these challenges, HAPFL employs a dual-agent DRL framework, which makes adaptive training adjustments by collaboratively leveraging an RL-based heterogeneous model allocation mechanism and an RL-based training intensity adjustment scheme. 
The former RL agent dynamically allocates different model sizes to clients based on their capabilities, aiming to mitigate performance discrepancies. The other RL agent further tailors the training intensities for each client, optimizing the balance between reducing straggling latency and ensuring sufficient training depth.

\subsubsection{RL-based Heterogeneous Model Allocation Mechanism}
To mitigate the significant performance differences among clients, one common intuitive strategy is to allocate larger, more capable models to clients with superior performance. This enables them to achieve better performance, enhanced generalization capabilities, and the ability to handle more complex tasks. Building on this strategy, the HAPFL method incorporates an RL-based heterogeneous model allocation mechanism using the PPO model, abbreviated as \textbf{PPO1}.
PPO1 utilizes a multi-head RL model architecture specifically designed for model allocation, enabling the mechanism to dynamically adjust model sizes based on each client's capabilities. The primary goal is to minimize performance disparities among clients and avoid the potential imbalances that may arise from varying training intensities alone.

\textbf{State space}: 
In each round of federated training, the server collects the training time costs $T^d_{r,i}$ from the performance assessments of participating clients. This data, utilized as the input for the PPO1 model, can be expressed as follows:
\begin{equation}
\label{eq:startStateOfPPO1}
    T^d_r = \{T^d_{r,i} | i \in \mathrm{S}_r\}.
\end{equation}
To effectively capture performance differences among clients, the collected assessment training time costs are normalized as:
\begin{equation}
    T'^d_{r,i} = \frac{T^d_{r,i}}{min(T^d_r)}.
\end{equation}
Thus, the state space $\mathcal{S}^1_r$ of PPO1 in round $r$ is defined as:
\begin{equation}
    \mathcal{S}^1_r = \{T'^d_{r,i} | i \in \mathrm{S}_r\}.
\end{equation}

\textbf{Action space}: The primary objective of PPO1 for model allocation is to balance performance differences among clients by dynamically assigning models of varying sizes. As such, the output of PPO1 determines the specific model category, reflecting different sizes, allocated to each participating client. The action space $\mathcal{A}^1_r$ of PPO1 for this decision-making process in round $r$ can be expressed as follows:
\begin{equation}
    \mathcal{A}^1_r=\{ a^1_{r,i}|i \in \mathrm{S}_r \},
\end{equation}
\begin{equation}
\label{eq:endActionOfPPO1}
    a^1_{r,i} = \{1, 2,\ldots, \delta \},
\end{equation}
where $ a^1_{r,i} $ represents the model category assigned to the respective client $i$ in the $r$-th round, and $ \delta $ represents the total number of model categories. Each category corresponds to a different model size, allowing PPO1 to facilitate heterogeneous training according to the individual capabilities and needs of each client, thereby promoting a more balanced and efficient learning process.

\textbf{Reward function}: 
Upon completion of the $r$-th round of federated training, the server collects the local training time costs $T^l_r$ from all participating clients $i$ within the set $\mathrm{S}_r$. This collection of training time costs can be formally expressed as:
\begin{equation}
\label{eq:startRewardOfPPO1}
    T^l_r=\{ T^l_{r,i} | i \in \mathrm{S}_r \}.
\end{equation}
The average training time $T^{l,avg}_{r,i}$ required for each participating client to complete one epoch is calculated based on the local training time $T^l_{r,i}$ and the number of iterations $\tau_{r,i}$:
\begin{equation}
    T^{l,avg}_{r,i} = \frac{T^l_{r,i}}{\tau_{r,i}},
\end{equation}
\begin{equation}
    T^{l,avg}_r = \{ T^{l,avg}_{r,i} | i \in \mathrm{S}_r \}.
\end{equation}
The reward function for the PPO1 model is then formulated to minimize the relative performance discrepancies among clients,  encouraging a more balanced training:
\begin{equation}
\label{eq:endRewardOfPPO1}
    \mathcal{R}^1_r = MD - \frac{max(T^{l,avg}_r)}{min(T^{l,avg}_r)},
\end{equation}
where $MD$ represents the maximum acceptable multiple of performance differences between participating clients after balancing the training with differentiated model allocations. This function aims to narrow the gap between the fastest and slowest clients, thus enhancing overall training efficiency.

\subsubsection{RL-based training intensity adjustment scheme}
Building on the achievements of PPO1 in balancing performance disparities among clients, the HAPFL method incorporates a second PPO-based decision-making scheme, referred to as \textbf{PPO2}.
This scheme is designed to dynamically assign varying training intensities to clients according to their performance levels, aiming to further optimize overall straggling latency and achieve adaptive training adjustments.
This dual-layered approach of PPO1 and PPO2 significantly boosts the efficiency and effectiveness of federated learning across heterogeneous client environments.

\textbf{State space}: 
The state space for PPO2 is formulated by integrating the normalized assessment training time $T'^d_{r,i}$ with the outcomes $a_{r,i}$ of PPO1. This integration can be expressed as follows:
\begin{equation}
\label{eq:startStateOfPPO2}
    T^m_{r,i} = M(a_{r,i})\ T'^d_{r,i},
\end{equation}
where $T^m_{r,i}$ represents the modified training time for client $i$ in round $r$, adjusted according to the model type allocated. $ M(\cdot) $ represents the relative training time ratio for local training corresponding to different types of models.
The collective state space $\mathcal{S}^2_r$ for PPO2 in round $r$ is then defined as:
\begin{equation}
    \mathcal{S}^2_r = \{ T^m_{r,i} | i \in \mathrm{S}_r \}.
\end{equation}

\textbf{Action space}: 
PPO2 aims to dynamically adjust training intensities according to the varying performance levels of clients to further optimize straggling latency. Thus, the action space $\mathcal{A}^2$ of PPO2 encompasses the decision regarding an array of normalized training intensities $[\sigma_i]_{1 \times k}$, derived through a Softmax normalization process, which are assigned to $k$ participating clients in round $r$:
\begin{equation}\centering
\begin{aligned}
    &\mathcal{A}^2_r = [\sigma_i]_{1 \times k}, \\
    0 \leqslant \sigma_i \leqslant 1, &\sum_{i\in [1,k]} \sigma_i=1, k=|\mathrm{S}_r|.
\end{aligned}
\end{equation}
Finally, the number of local training iterations $\tau_{r,i}$ assigned to client $i$ in round $r$ is calculated by multiplying the normalized array $[\sigma_i]_{1 \times k}$ by the total training intensity $\tilde{\tau}_r$, as follows:
\begin{equation}
    [\tau_{r,i}]_{1 \times k} = [\sigma_i]_{1 \times k} * \tilde{\tau}_r.
\end{equation}



\textbf{Reward function}: The server receives the local training time, denoted as $ T^l_r = \{T^l_{r,i}\},i\in \mathrm{S}_r$, from each participating client and computes the straggling latency for the current round. The reward function for the RL process is defined as:
\begin{equation}
\label{eq:rewardOfPPO2}
    \mathcal{R}^2_r = min(T^l_r) - max(T^l_r).
\end{equation}

\subsubsection{RL Learning Process}
The Proximal Policy Optimization \cite{schulman2017proximal} algorithm stands out for its efficiency, robust convergence properties, and suitability for continuous action spaces. In this research, we leverage the PPO algorithm to train the above two RL models. The PPO algorithm maintains an actor network $ \pi_{\theta}(\mathcal{A}_r|\mathcal{S}_r) $  and a critic network  $ V_{\phi}(\mathcal{S}_r) $, where $ \theta $ and $ \phi $ represent the parameters of the actor and critic networks, respectively. 
To further enhance the learning process, both PPO1 and PPO2 models incorporate an experience replay buffer that retains a history of past states, actions, rewards, enabling the RL model to learn from a diverse range of experiences and improve the robustness of the policy.


The objective of our DRL agents is to maximize the cumulative discounted return,  $ G_r $, defined as:
\begin{equation}
\label{eq:startUpdatePPO}
    G_r = \sum_{t=0}^{\infty} \gamma^t \mathcal{R}_{r+t},
\end{equation}
where $G_r$ indicates the cumulative discounted return at round $r$, $0 \leq \gamma \leq 1$ is the discount factor that prioritizes immediate rewards over distant ones, and $\mathcal{R}_{r+t}$ represents the reward at round $r+t$.

\textbf{Actor Network}: The actor network is designed to output the probability distribution of potential actions for a given state. It plays a crucial role in selecting actions based on the current state by implementing the parameterized policy function. 
During training, the primary objective of the actor network is to maximize the expected reward, which is achieved by adjusting the probabilities to favor actions that increase the expected returns under the current policy.
The parameters of the actor network are updated using the policy gradient method to progressively converge the output action probabilities toward the optimal value. The loss function for the actor network is defined as follows:
\begin{equation}
    L(\theta) = \mathbb{E} \left[ \min \left( \rho_r(\theta) \widehat{A}_r, \text{clip} \left( \rho_r(\theta), 1-\epsilon, 1+\epsilon \right) \widehat{A}_r \right) \right],
\end{equation}
where $ \theta $ denotes the parameters of the actor network, $ \rho_r(\theta) $ represents the action probability ratio reflecting the trust region, $ \widehat{A}_r $ is the advantage function, $ \epsilon $ is a hyperparameter that helps limit the policy update step, and ``clip" refers to a clip function. The advantage function is given by:
\begin{equation}
    \widehat{A}_r = G_r - V(\mathcal{S}_r),
\end{equation}
where $ G_r $ represents the cumulative discounted return, and $ V(\mathcal{S}_r) $ is the value estimate provided by the critic network for the state $ \mathcal{S}_r $. 

\textbf{Critic network}: The critic network is responsible for estimating the value of the current state, providing an evaluation of the quality of each state to guide the action selection and optimization processes.
The primary goal of the critic network is to accurately approximate the true state value, which involves minimizing the distance between the estimated and true state values.
The parameters of the critic network are iteratively updated based on the value function loss, which quantifies the accuracy of the critic's predictions and facilitates the convergence of the estimated values towards their true values. The loss function for the critic network is defined as:
\begin{equation}
\label{eq:endUpdatePPO}
    L(\phi) = \mathbb{E} \left[ \left( V(\mathcal{S}_r) - G_r \right)^2 \right],
\end{equation}
where $ \phi $ represents the parameters of the critic network, $ V(\mathcal{S}_r) $ is the estimated value of the state $ \mathcal{S}_r $, and $ G_r $ represents the cumulative discounted return.

\subsection{Local Training}
The primary goal of local training in HAPFL is to optimize the use of local data for enhancing model performance while protecting data privacy. To achieve this, in our HAPFL method, the server adaptively allocates appropriate heterogeneous models to clients based on their varying computing capabilities and then aggregates these heterogeneous models on the server, respectively. Nevertheless, this respective aggregation of heterogeneous models can restrict clients from fully leveraging the knowledge pooled from all participants.
To address this limitation, the HAPFL method employs a knowledge distillation-based mutual learning mechanism for local training. In this mechanism, each client operates with two models: a globally unified architecture LiteModel that aggregates and distills global knowledge and a heterogeneous local model tailored to specific client capabilities. 
The training of the local model is guided by the LiteModel to effectively incorporate global insights, while the local model enriches the LiteModel with specific local knowledge. 

\textbf{Local Model Training}: To effectively harness global knowledge, the local model undergoes training using a knowledge distillation approach, with its loss function defined as:
\begin{equation}
\label{eq:localModelLossFunction}
    L_1=-\frac{\lambda_1}{N}\sum^N_{i=1}\sum^M_{j=1}y_{ij}log(\hat{y}_{ij}) + \lambda_2 D_{KL}(X \left| \right| Y),
\end{equation}
where the first term is the cross-entropy loss function, $ N $ represents the number of samples in the dataset, and $ M $ represents the number of categories in each sample. $ y_{ij} $ represents the actual label of the $ j $-th category of the $ i $-th sample.
$ \hat{y}_{ij} $ is the model's predicted value for the $ j $-th category of the $ i $-th sample.
The second term is the relative entropy loss function, $ D_{KL}(X \left| \right| Y) $ denotes the Kullback-Leibler (KL) divergence between the true distribution $ X $ and the model's distribution $ Y $, $ X $ is the output of the local model, and $ Y $ is the output of the LiteModel. $ \lambda_1 $ and $ \lambda_2 $ are hyperparameters that control the weight of cross-entropy and KL divergence, respectively, with $ \lambda_1+\lambda_2=1 $.

\textbf{LiteModel Training}: The LiteModel is a globally unified model whose role is to aggregate and relay global knowledge to assist local training.
Its loss function is similarly structured to encourage consistency between global and local predictions:
\begin{equation}
\label{eq:litemodelLossFunction}
    L_2=-\frac{\lambda_3}{N}\sum^N_{i=1}\sum^M_{j=1}y_{ij}log(\hat{y}_{ij}) + \lambda_4 D_{KL}(Y \left| \right| X),
\end{equation}
where $ \lambda_3 + \lambda_4 = 1$ ensuring a balanced adjustment between direct learning and divergence minimization.

The model is updated using the stochastic gradient descent algorithm, with the update expression given by:
\begin{equation}
\label{eq:gradientUpdate}
    \theta_{i}^{r+1} = \theta^r_i - \eta \nabla L,
\end{equation}
where $ \theta^r_i $ represents the model parameters of client $ i $ in the current round $ r $, $ \theta^{r+1}_i $ represents the new model parameters of participant $ i $ after local update, $ \eta $ denotes the learning rate, $ \nabla L $ denotes the gradient of the loss function $ L $ with respect to the model parameter $ \theta^r_i $.

This dual-model structured approach ensures that both local and global models benefit from continuous, bidirectional learning, optimizing both specific and general performance across the federated network.

\subsection{Global Model Aggregation}
The commonly employed aggregation method in Federated Learning is FedAvg \cite{mcmahan2017communication}. However, FedAvg often underperforms in non-IID settings. Specifically, due to disparities in local data distributions, model parameters trained by different participants may exhibit significant variance. Such discrepancies necessitate more communication rounds for convergence during the model aggregation process, thereby diminishing the efficiency of federated learning. Moreover, these variances can result in suboptimal model performance across various participants, leading to poor overall model performance post-aggregation. This challenge is further exacerbated in environments where training intensities vary among clients, for which the HAPFL method adopts a sophisticated weighted aggregation approach.

Information entropy \cite{shannon1948mathematical} is a metric that quantifies the uncertainty or the amount of information contained within a dataset.
The information entropy of a dataset can be calculated by first determining the frequency of occurrence of each category within the dataset.
Assuming a dataset comprises $ n $ samples across $ C $ different categories, with $ M_i $ denoting the number of samples in category $ i $, the probability of category $ i $ is given by:
\begin{equation}
\label{eq:startOfAggregationWeight}
    Q(i)=\frac{M_i}{n}, i \in C.
\end{equation}
The information entropy $ H $ of the dataset is then calculated by the following formula:
\begin{equation}
    H = - \sum^K_{i=1}Q(i)log_2 Q(i), i \in C,
\end{equation}
where a higher $ H $ indicates greater dataset uncertainty and more information, whereas a lower $ H $ signifies lower uncertainty and less information.

In the HAPFL method, information entropy and model accuracy are combined to define the aggregation weights for different clients, enhancing the fairness and effectiveness of the model aggregation:
\begin{equation}
\label{eq:endOfAggregationWeight}
    W_r = \frac{1}{2} (softmax(H_r)+softmax(acc_r)),
\end{equation}
where $ W_r $ represents the aggregation weight for the participating clients in round $ r $, $ H_r $ is the information entropy, and $ acc_r $ indicates the model's accuracy after local training in round $ r $.

Note that the server conducts separate aggregation processes for homogeneous LiteModels and heterogeneous local models collected from the clients, where the same-sized local models are aggregated together. All these aggregated global models are updated as follows:
\begin{equation}
\label{eq:globalModelAggregation}
    \theta^{global}_{r+1} = \theta^{global}_{r} + \sum W_{r,i} \cdot \theta_{r,i},i \in \mathrm{S}_r,
\end{equation}
where $\theta^{global}_{r}$ denotes the global model parameters in round $r$, and $ \theta_{r,i} $ indicates each client's model parameters that are weighted by $W_{r,i}$, a factor derived from both the information entropy and model accuracy.
This aggregation scheme ensures that the global model not only accumulates knowledge uniformly but also respects each client's unique contributions based on the diversity of their data and their model's performance, aiming to produce a more robust and effective federated model.

\subsection{Convergence Analysis}
\textbf{Convergence Analysis in Convex Settings}:In this section, we delve into the convergence properties of the objective function denoted by:
\begin{equation}
\label{eq:objectivefunction}
    \mathcal{L}(x) = \mathcal{L}_{CE}(x)+\lambda \mathcal{L}_{KL}(x),
\end{equation}
where $ \mathcal{L}_{CE}(x) $ is cross-entropy loss, $ \mathcal{L}_{KL}(x) $ is KL divergence term, and $ \lambda $ is a weighting factor. Our analysis begins with the following assumptions.

\textbf{Assumption 1.} ($ L-Lipschitz \ smoothness $) The function $ f $ has an $ L-Lipschitz $ continuous gradient, i.e., there exists a constant $ L>0 $ such that for all $ x,y \in \mathbb{R}^n $, the following holds: $ \Vert \nabla f(x) - \nabla f(y) \Vert \leq L \Vert x - y \Vert $.

\textbf{Assumption 2.} ($\mu$-$strongly$) The function $ f $ is $\mu$-$strongly$ convex, signifying there exists a constant $ \mu > 0 $, such that for all $ x,y \in \mathbb{R}^n $ , the following holds: $ f(y) \geq f(x) + \nabla f(x)^T(y-x) + \frac{\mu}{2}\Vert y - x \Vert^2 $.

From Assumption 1 and Assumption 2, it can be inferred that the composite loss function $ \mathcal{L}(x) $ is strongly convex and the gradient of $ \mathcal{L}(x) $ is $ Lipschitz $ continuous, where the $ Lipschitz $ constant $ L $ can be expressed as $ L=L_{CE}+\lambda L_{KL} $.

The iterative update rule for the gradient descent algorithm is given by:
\begin{equation}
    x_{\kappa+1}=x_\kappa-\eta\nabla \mathcal{L}(x_\kappa).
\end{equation}

Under the assumption of $L-Lipschitz$ smoothness, the convergence of the gradient descent for the composite loss function $ \mathcal{L}(x) $ can be analogous to the convergence of a single loss function. Based on assumption 1, consider the following inequality:
\begin{equation}
    \mathcal{L}(x_{\kappa+1}) \leq\mathcal{L}(x_\kappa) + \nabla \mathcal{L}(x_\kappa)^T(x_{\kappa+1}-x_\kappa)+\frac{L}{2} \Vert x_{\kappa+1} - x_\kappa \Vert^2.
\end{equation}
Substituting the gradient descent update rule, we can obtain:
\begin{equation}
\label{eq:non-convex}
    \mathcal{L}(x_{\kappa+1}) \leq \mathcal{L}(x_\kappa) - (\eta - \frac{L\eta^2}{2}) \Vert \nabla \mathcal{L}(x_\kappa)\Vert^2.
\end{equation}
This implies that the decrease in the loss function value at each iteration is proportional to the square of the gradient norm, exhibiting sublinear convergence. According to Assumption 2, we can obtain:
\begin{equation}
    \mathcal{L}(x)-\mathcal{L}^* \geq \frac{\mu}{2}\Vert x - x^*\Vert^2,
\end{equation}
where $\mathcal{L}^*$ is the loss value corresponding to the global optimal solution. Combining with the gradient descent update rule, we can obtain:
\begin{equation}
    \Vert \nabla \mathcal{L}(x_\kappa) \Vert^2 \geq 2 \mu (\mathcal{L}(x_\kappa) - \mathcal{L}^*).
\end{equation}
Substituting it into the derivation of $Lipschitz$ continuity, we can obtain:
\begin{equation}
    \mathcal{L}(x_{\kappa+1}) \leq \mathcal{L}(x_\kappa)-2\mu(\eta-\frac{L\eta^2}{2})(\mathcal{L}(x_\kappa)-\mathcal{L}^*).
\end{equation}
Let $\Gamma = 2\mu(\eta - \frac{L\eta^2}{2})$, rearranging, we can obtain:
\begin{equation}
    \mathcal{L}(x_{\kappa+1}) - \mathcal{L}^* \leq (1-\Gamma)(\mathcal{L}(x_\kappa)-\mathcal{L}^*).
\end{equation}
Through recursive iteration, we can obtain:
\begin{equation}
    \mathcal{L}(x_\kappa)-\mathcal{L}^* \leq (1-\Gamma)^{\kappa}(\mathcal{L}(x_0)-\mathcal{L}^*).
\end{equation}
As $\kappa$ tends to infinity, if the learning rate $\eta$ is chosen small enough such that $0 < \eta\mu < 2$, then the convergence rate is:
\begin{equation}
    \lim_{t \to \infty}(1-\Gamma)^{\kappa}=0.
\end{equation}

This indicates that gradient descent converges to the global optimum at a linear rate of $O((1-\Gamma)^\kappa)$.

\textbf{Convergence Analysis in Non-Convex Settings}:
In the non-convex setting, we analyze the convergence of the gradient descent method for the objective function (\ref{eq:objectivefunction}). According to Assumption 1, the gradient of the objective function satisfies $L-Lipschitz \ smoothness$, so the above formula (\ref{eq:non-convex}) can be derived.

\textbf{Non-Convex Convergence Guarantee.} Define $\Delta_\mathcal{L} = \mathcal{L}(x_0) - \mathcal{L}{\inf}$, where $\mathcal{L}{\inf}$ is the infimum of $\mathcal{L}(x)$. Summing the above inequality over $\kappa = 0, \ldots, K-1$, we have:

\begin{equation}
    \Delta_\mathcal{L} \geq \sum_{\kappa=0}^{K-1} \left(\eta - \frac{L\eta^2}{2}\right) \|\nabla \mathcal{L}(x_\kappa)\|^2.
\end{equation}

Let $\eta$ be chosen such that $\eta < \frac{2}{L}$ to ensure $\eta - \frac{L\eta^2}{2} > 0$. Then:

\begin{equation}
    \frac{1}{K} \sum_{\kappa=0}^{K-1} \|\nabla \mathcal{L}(x_\kappa)\|^2 \leq \frac{\Delta_\mathcal{L}}{K \left(\eta - \frac{L\eta^2}{2}\right)}.
\end{equation}

As $K \to \infty$, the average gradient norm converges to zero:

\begin{equation}
    \lim_{K \to \infty} \frac{1}{K} \sum_{\kappa=0}^{K-1} \|\nabla \mathcal{L}(x_\kappa)\|^2 = 0.
\end{equation}

For a non-convex objective function, gradient descent ensures convergence to a stationary point (where $|\nabla \mathcal{L}(x)| \to 0$) at a sublinear rate of $O(1/K)$ in terms of the average gradient norm.

\section{Performance Evaluation}
\label{evaluation}
In this section, we conducted extensive simulations to validate the performance of our proposed HAPFL method. The experimental results demonstrate that HAPFL is superior to its competitors in terms of training efficiency and model accuracy.

\subsection{Experiment Settings}
\subsubsection{Testbed Setup}
To evaluate the effectiveness of the HAPFL, we implemented this framework using PyTorch (version 1.12.1) and conducted a series of simulations on a high-performance server. The server is equipped with an Intel Xeon Silver 4314 CPU @ 2.40GHz, 192GB of memory, and an NVIDIA GeForce GTX 3080 GPU, running the Ubuntu 5.4.0-182-generic operating system.

\subsubsection{Dataset Settings}
Our experiments utilized three widely recognized image datasets, i.e., MNIST, CIFAR-10, and ImageNet-10, to ensure a comprehensive evaluation across varying degrees of image complexity. 
\begin{itemize}
    \item MNIST: This dataset comprises handwritten digit images categorized into 10 classes (0-9), with images characterized by simplicity and clarity. The dataset contains 60,000 training images and 10,000 test images, each being grayscale and having a resolution of 28x28 pixels.
    \item CIFAR-10: This dataset consists of 10 classes (e.g., airplanes, cars, birds) with approximately 5,000 training images per class. These images exhibit moderate complexity, variations, and background noise. CIFAR-10 comprises 50,000 training images and 10,000 test images, each being a color image with a resolution of 32x32 pixels.
    \item ImageNet-10: This dataset, derived from the larger ImageNet dataset, includes 7,200 training images and 1,800 test images, featuring high complexity and diverse scenes. To maximize data utilization, we expanded the dataset by employing techniques such as random cropping and random horizontal flipping.
\end{itemize}

\subsubsection{Model Settings}
In our experimental setup, the HAPFL framework consists of two functional RL models and three types of FL models.
\begin{itemize}
    \item RL Models: Two RL models are employed in our approach, one for model allocation and the other for training intensity allocation. Each RL model employs PPO, comprising an actor network and a critic network. The model allocation PPO1 utilizes a multi-head neural network as the actor network and a three-layer fully connected neural network as the critic network. The training intensity allocation PPO2 employs a four-layer fully connected neural network for both actor network and critic network. 
    \item FL Models: We configure three types of FL models: a LiteModel, a small model, and a large model. Each client holds two models: a LiteModel used to ensure consistency in learning and data synthesis, and a local model with different sizes (i.e., small, and large) depending on the client's computing capacity. These FL models are built using Convolutional Neural Networks (CNNs) \cite{krizhevsky2012imagenet1} tailored to different datasets.
\end{itemize}

\subsubsection{FL Settings}
In our experiments, we simulated an FL scenario consisting of 10 heterogeneous clients with varied computational resources and data distributions, where the test images from different datasets are allocated to these clients using Dirichlet partitioning. Table 1 outlines the key hyperparameters used in the experiments.

\begin{table}[!htbp]
\caption{List of hyperparameters} 
\label{tab:tableTab} 
\centering
\begin{tabular}{cc} 
\hline 
Parameters & Values \\
\hline 
Total number of clients, $ K $ & 10 \\
Randomly selected clients per round, $ k $ & 6 \\
Maximum performance difference between clients, $ MD $ &10 \\
Loss function weight, $ \lambda_1 $ & 0.4 \\
Loss function weight, $ \lambda_2 $ & 0.6 \\
Loss function weight, $ \lambda_3 $ & 0.5 \\
Loss function weight, $ \lambda_4 $ & 0.5 \\
Truncation function hyperparameter, $ \epsilon $ & 0.2 \\
Learning rate of PPO1, $ lr_1 $ & 0.02 \\
Learning rate of PPO2, $ lr_2 $ & 0.0003 \\
Batch size to update RL agents, $ B $ & 5 \\
FL model learning rate, $ lr_3 $ & 0.0003 \\
Dirichlet partition rate, $ \alpha $ & 0.4 \\
Default number of local epochs, $ E $ & 20 \\
\hline 
\end{tabular}
\end{table}

\subsection{Compared Algorithms}
To demonstrate the effectiveness and superiority of HAPFL, we compared it with the following three baseline algorithms.
\begin{enumerate}
    \item FedAvg \cite{mcmahan2017communication}: FedAvg is a foundational FL algorithm that aggregates model updates by averaging the parameters updated locally by clients. This method assumes a uniform model architecture across all clients and applies the same training intensity for local training.
    \item FedProx \cite{li2020federated}: 
    FedProx is an advanced FL algorithm designed to better accommodate challenges in federated settings, such as non-convex optimization tasks and unbalanced data distributions among clients. The key principle of the FedProx algorithm is the integration of a proximal term to moderate the difference between local updates and the global model. This effectively enhances the alignment between local and global model performance with more consistent and stable convergence.

    \item pFedMe \cite{t2020personalized}: pFedMe is a personalized federated learning algorithm designed to enhance the learning performance of clients by leveraging the advantages of both personalized and global models. The key principle of the pFedMe algorithm is to allow each client to perform personalized training on its local model while introducing a regularization term to constrain the deviation between the local model and the global model. This approach effectively facilitates personalized adjustments among clients, improving model performance in heterogeneous data environments. By considering personalized updates during the aggregation process, pFedMe enables the global model to better adapt to the data distributions of different clients, resulting in higher accuracy and stability.

    \item FedDdrl\cite{wong2023fedddrl}: FedDdrl a federated learning approach designed to address device heterogeneity in IoT environments. It uses a double deep reinforcement learning mechanism to dynamically adjust local training epochs and adopt early termination strategies, effectively reducing training latency and mitigating the straggler problem caused by varying client performance. This method optimizes resource allocation to improve training efficiency while accommodating resource-constrained devices.
\end{enumerate}

\subsection{Experimental Results}
\subsubsection{Performance of Dual-Agent RL Models}
\begin{figure}[htbp]
    \centering
    \begin{minipage}[b]{0.24\textwidth}
        \includegraphics[width=\textwidth]{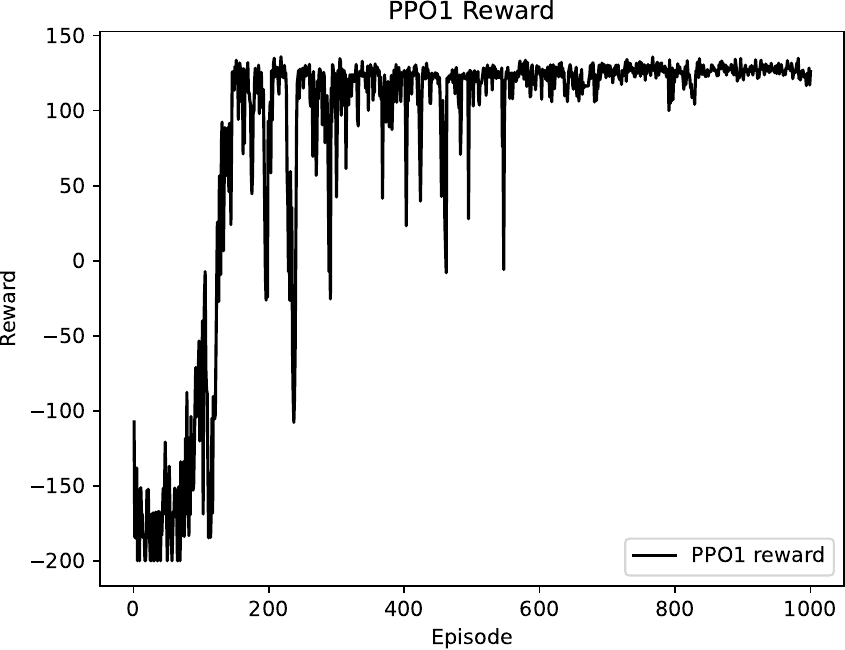}
        \caption{Reward for PPO1}
        \label{fig:Reward for PPO1}
    \end{minipage}
    \hfill 
    \begin{minipage}[b]{0.24\textwidth}
        \includegraphics[width=\textwidth]{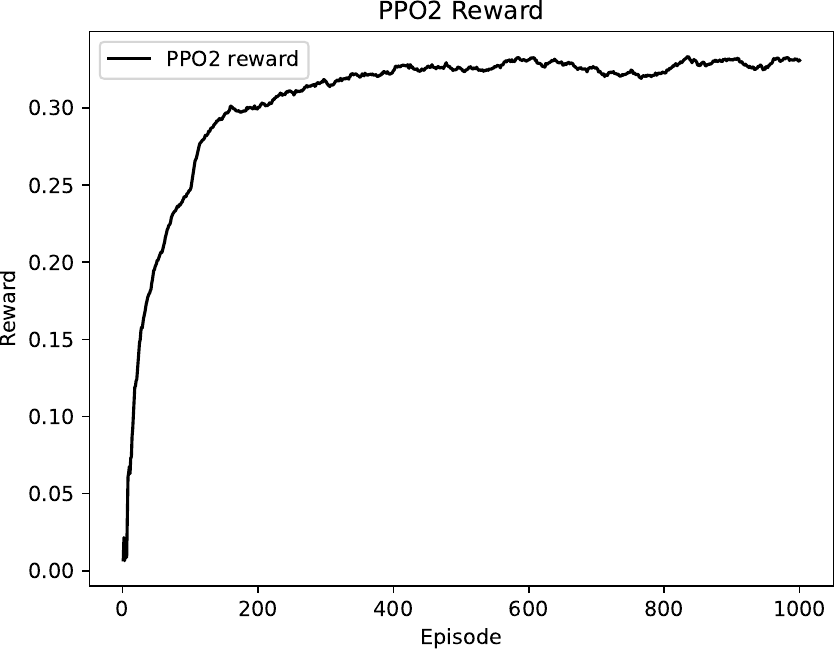}
        \caption{Reward for PPO2}
        \label{fig:Reward for PPO2}
    \end{minipage}
    \vspace{-0.3cm}
\end{figure}

In this evaluation, we focus on validating the performance of two customized PPO models, specifically designed to respond to dynamic changes in client performance within a federated learning environment. The effectiveness and adaptability of these models are crucial for the overall success of the HAPFL method.
Figure \ref{fig:Reward for PPO1} presents the trend of rewards for the PPO1 model across training iterations. It can be observed that the reward curve hovers around -180 in the early stages, indicating early adaptation changes. As training progresses, the curve experiences sharp fluctuations, highlighting the model's responsiveness to dynamic changes in client performance. Over time, these fluctuations diminish, and the rewards begin to show a gradual upward trend, though minor fluctuations persist.  This pattern suggests that the PPO1 model gradually adapts and stabilizes in response to dynamic client performance variability.
The performance trend for the PPO2 model, as depicted in Figure \ref{fig:Reward for PPO2}, shows a more consistent and smoother upward trend. Early in the training process, the rewards increase rapidly, demonstrating the model's effective initial adaptation. As training iterations continue, the rate of reward increase slows but maintains a steady ascent, suggesting that PPO2, with its specific adjustments for training intensity, manages to more effectively mitigate the impact of performance variations among clients over time.

These observations prove the distinct roles and effectiveness of each RL model within the HAPFL framework. PPO1, focusing on model allocation, shows robust adaptation, while PPO2, which adjusts training intensities, offers a more refined training adjustment to the dynamic environment, leading to further performance improvements. This highlights the complementary nature of the dual-agent approach in managing the complexities of dynamic heterogeneous FL environments.

\subsubsection{Performance of FL Models}
In this subsection, we evaluate the testing accuracy and loss of the HAPFL method through simulation experiments, comparing it with three baseline algorithms—FedAvg, FedProx, and pFedMe—across the MNIST, CIFAR-10, and ImageNet-10 datasets. For FedAvg and FedProx, accuracy and loss are assessed on all three datasets, while for pFedMe, we specifically compare the maximum and average accuracy of local models on MNIST and CIFAR-10. The detailed evaluation results are presented below.

\textbf{MNIST:} Figures \ref{fig:image1}-\ref{fig:loss_image3} illustrate the comparison of accuracy and loss for the LiteModel, small model, and large model under the HAPFL method against the baseline algorithms on the MNIST dataset. 
Figures \ref{fig:image1} and \ref{fig:loss_image1} indicate that the LiteModel, which aggregates global client data, performs nearly identically to the baselines in both accuracy and loss.
For the remaining figures, the HAPFL method demonstrates superior performance in terms of accuracy and loss compared with FedProx and FedAvg for both large and small models, substantially outperforming its competitors with enhanced stability across all models. 
Table \ref{tab:MNIST PFL model comparison} further supports this by presenting the maximum and average accuracy across two heterogeneously assigned models on 10 local clients, confirming that HAPFL consistently surpasses pFedMe in both metrics. Overall, the HAPFL models achieve consistently high accuracy, around 98.5\%, highlighting the method’s effectiveness in sustaining performance and stability across diverse FL environments.


\begin{figure}[htbp]
    \centering
    \begin{minipage}[b]{0.24\textwidth}
        \includegraphics[width=\textwidth]{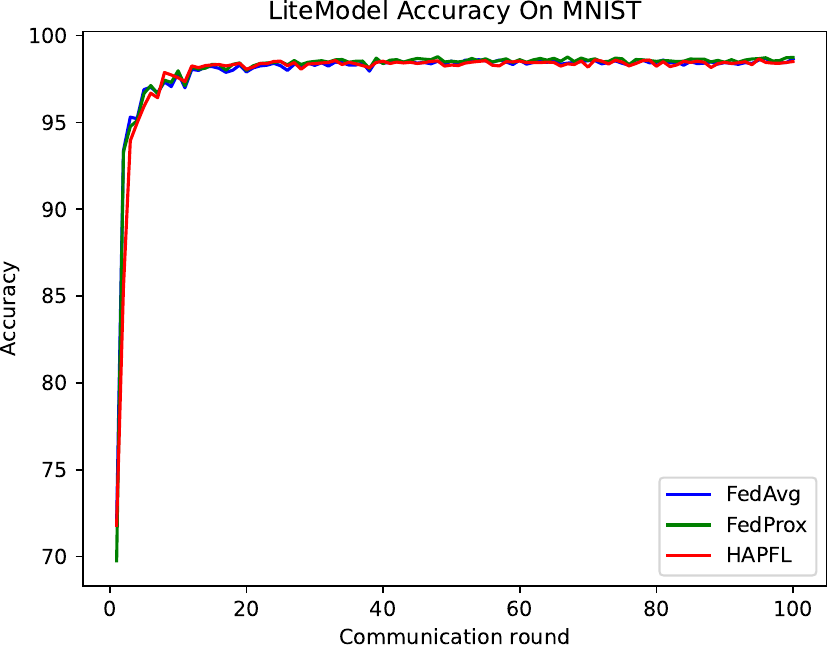}
        \caption{LiteModel on MNIST}
        \label{fig:image1}
    \end{minipage}
    \hfill 
    \begin{minipage}[b]{0.24\textwidth}
        \includegraphics[width=\textwidth]{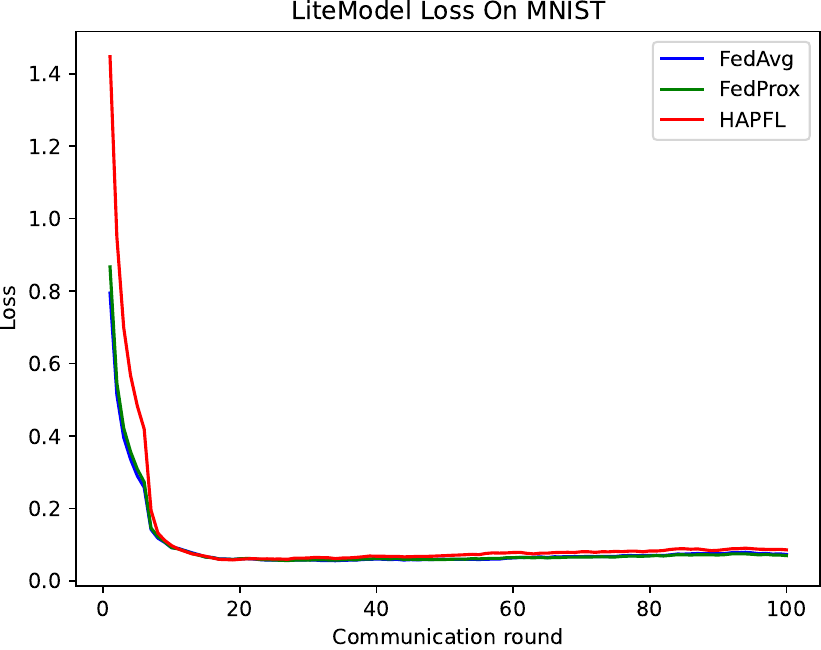}
        \caption{LiteModel on MNIST}
        \label{fig:loss_image1}
    \end{minipage}
    \vspace{-0.4cm}
\end{figure}

\begin{figure}[htbp]
    \centering
    \begin{minipage}[b]{0.24\textwidth}
        \includegraphics[width=\textwidth]{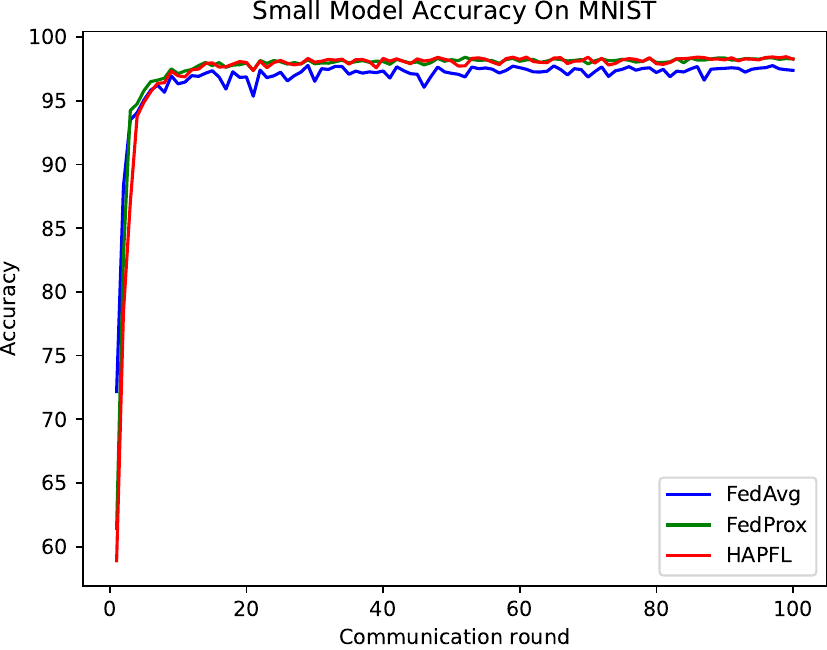}
        \caption{Small model on MNIST}
        \label{fig:image2}
    \end{minipage}
    \hfill 
    \begin{minipage}[b]{0.24\textwidth}
        \includegraphics[width=\textwidth]{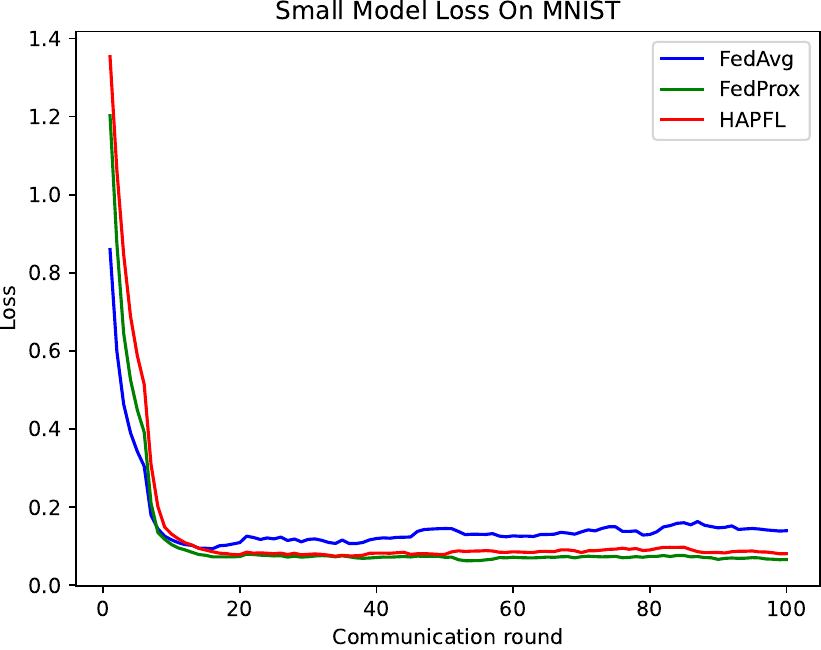}
        \caption{Small model on MNIST}
        \label{fig:loss_image2}
    \end{minipage}
    \vspace{-0.4cm}
\end{figure}

\begin{figure}[htbp]
    \centering
    \begin{minipage}[b]{0.24\textwidth}
        \includegraphics[width=\textwidth]{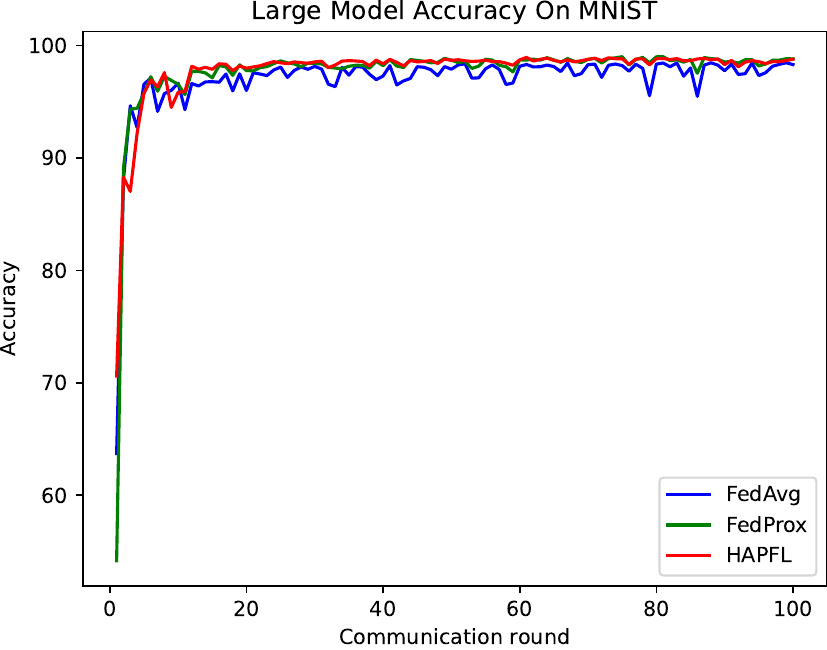}
        \caption{Large model on MNIST}
        \label{fig:image3}
    \end{minipage}
    \hfill 
    \begin{minipage}[b]{0.24\textwidth}
        \includegraphics[width=\textwidth]{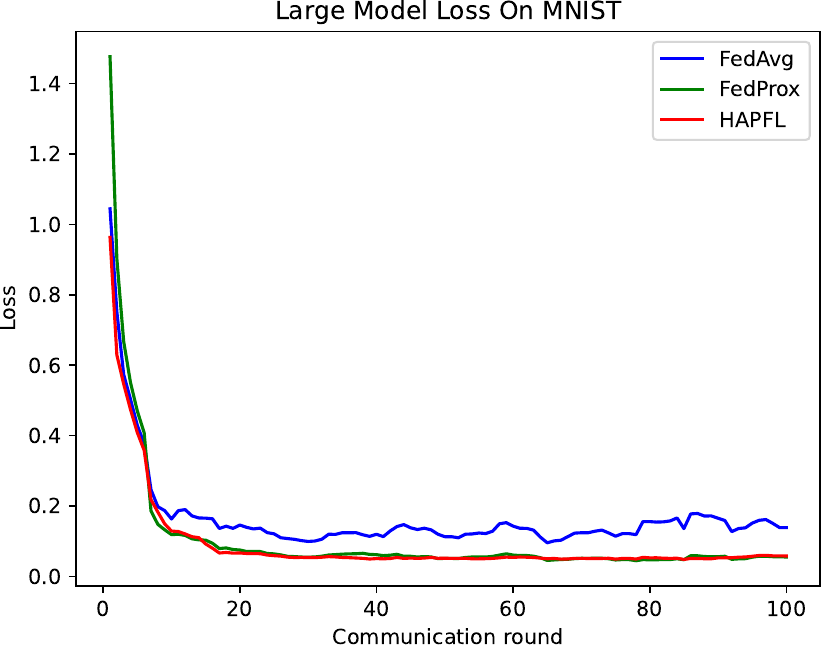}
        \caption{Large model on MNIST}
        \label{fig:loss_image3}
    \end{minipage}
    \vspace{-0.4cm}
\end{figure}


\begin{table}[htbp]
\centering
\caption{Comparison of personalized model accuracy between HAPFL and pFedMe algorithms on MNIST dataset}
\label{tab:MNIST PFL model comparison}
\resizebox{\linewidth}{!}{
\begin{tabular}{|c|c|c|c|c|c|}
\hline
\multirow{2}{*}{Client} & \multirow{2}{*}{Algorithm} & \multicolumn{2}{c|}{Small Model} & \multicolumn{2}{c|}{Big Model} \\ \cline{3-6} & & Maximum Accuracy & Average Accuracy & Maximum Accuracy & Average Accuracy \\ \hline
\multirow{2}{*}{Client 1} & HAPFL & 98.28\% & 97.90\% & 98.58\% & 97.74\% \\ \cline{2-6} & pFedMe & 89.88\% & 84.42\% & 80.32\% & 79.83\% \\ \hline
\multirow{2}{*}{Client 2} & HAPFL & 97.86\% & 95.96\% & 97.06\% & 97.06\% \\ \cline{2-6} & pFedMe & 87.64\% & 85.90\% & 91.25\% & 91.25\% \\ \hline
\multirow{2}{*}{Client 3} & HAPFL & - & - & 98.01\% & 95.59\% \\ \cline{2-6} & pFedMe & - & - & 74.46\% & 73.75\% \\ \hline
\multirow{2}{*}{Client 4} & HAPFL & 98.33\% & 96.73\% & - & - \\ \cline{2-6} & pFedMe & 84.83\% & 82.50\% & - & - \\ \hline
\multirow{2}{*}{Client 5} & HAPFL & 98.29\% & 96.97\% & - & - \\ \cline{2-6} & pFedMe & 89.79\% & 89.05\% & - & - \\ \hline
\multirow{2}{*}{Client 6} & HAPFL & - & - & 98.11\% & 94.22\% \\ \cline{2-6} & pFedMe & - & - & 65.23\% & 60.68\% \\ \hline
\multirow{2}{*}{Client 7} & HAPFL & 98.15\% & 96.45\% & 98.29\% & 96.83\% \\ \cline{2-6} & pFedMe & 86.86\% & 86.55\% & 87.36\% & 87.04\% \\ \hline
\multirow{2}{*}{Client 8} & HAPFL & 89.31\% & 89.31\% & 98.35\% & 96.57\% \\ \cline{2-6} & pFedMe & 87.22\% & 87.22\% & 80.54\% & 78.96\% \\ \hline
\multirow{2}{*}{Client 9} & HAPFL & 98.16\% & 96.20\% & - & - \\ \cline{2-6} & pFedMe & 75.56\% & 73.47\% & - & - \\ \hline
\multirow{2}{*}{Client 10} & HAPFL & 96.11\% & 96.11\% & 97.86\% & 94.87\% \\ \cline{2-6} & pFedMe & 93.12\% & 93.12\% & 81.92\% & 80.65\% \\ \hline
\end{tabular}}
\end{table}

\begin{figure}[htbp]
    \centering
    \begin{minipage}[b]{0.24\textwidth}
        \includegraphics[width=\textwidth]{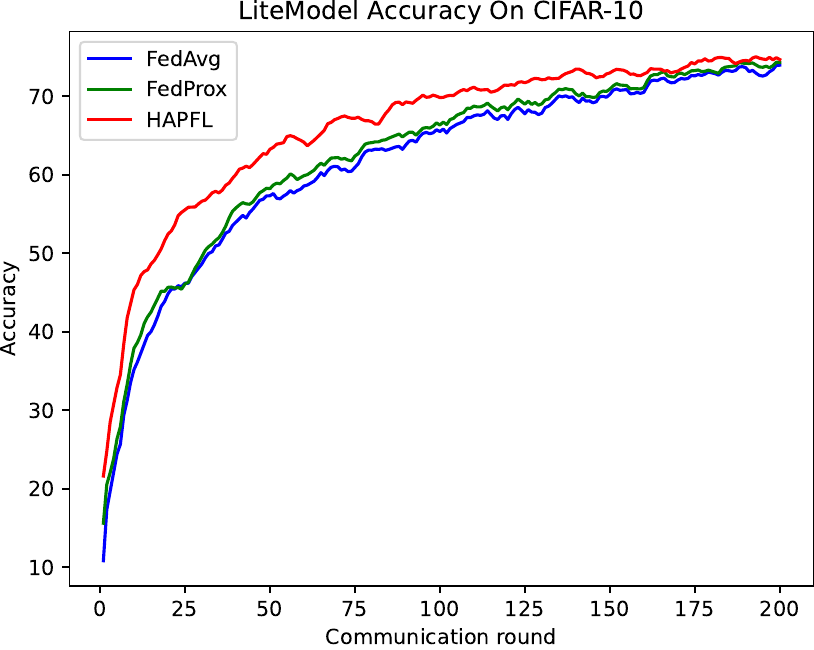}
        \caption{LiteModel on CIFAR-10}
        \label{fig:image4}
    \end{minipage}
    \hfill 
    \begin{minipage}[b]{0.24\textwidth}
        \includegraphics[width=\textwidth]{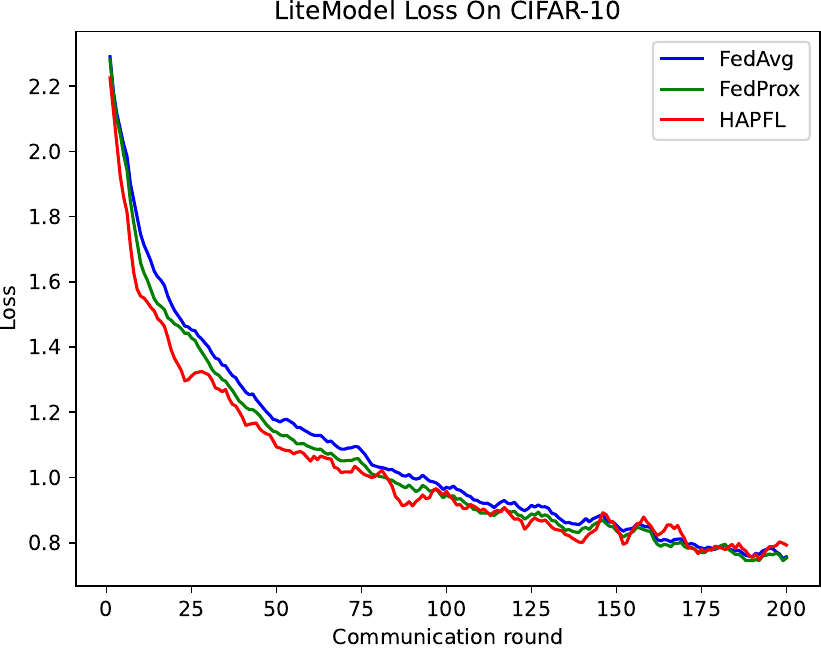}
        \caption{LiteModel on CIFAR-10}
        \label{fig:loss_image4}
    \end{minipage}
    \vspace{-0.4cm}
\end{figure}

\begin{figure}[htbp]
    \centering
    \begin{minipage}[b]{0.24\textwidth}
        \includegraphics[width=\textwidth]{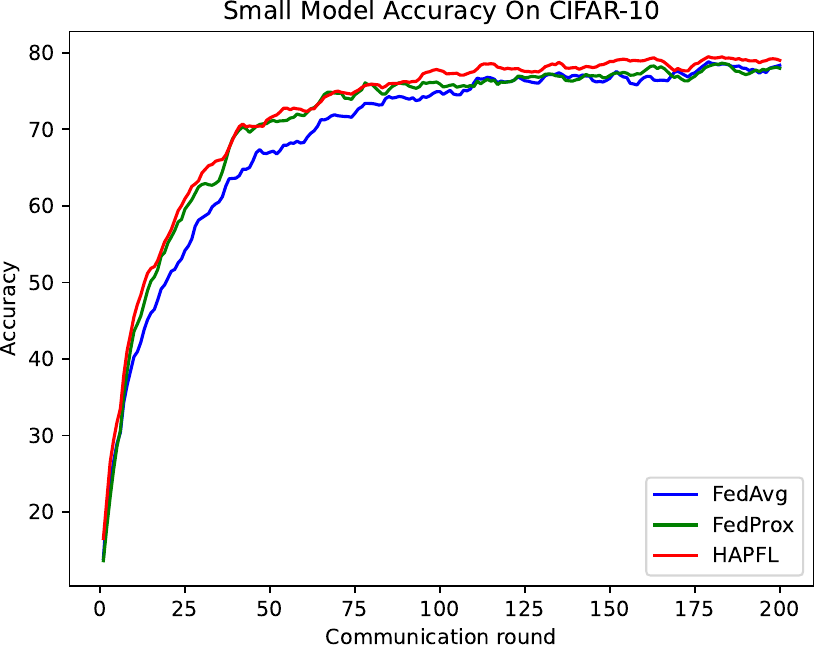}
        \caption{Small model on CIFAR-10}
        \label{fig:image5}
    \end{minipage}
    \hfill 
    \begin{minipage}[b]{0.24\textwidth}
        \includegraphics[width=\textwidth]{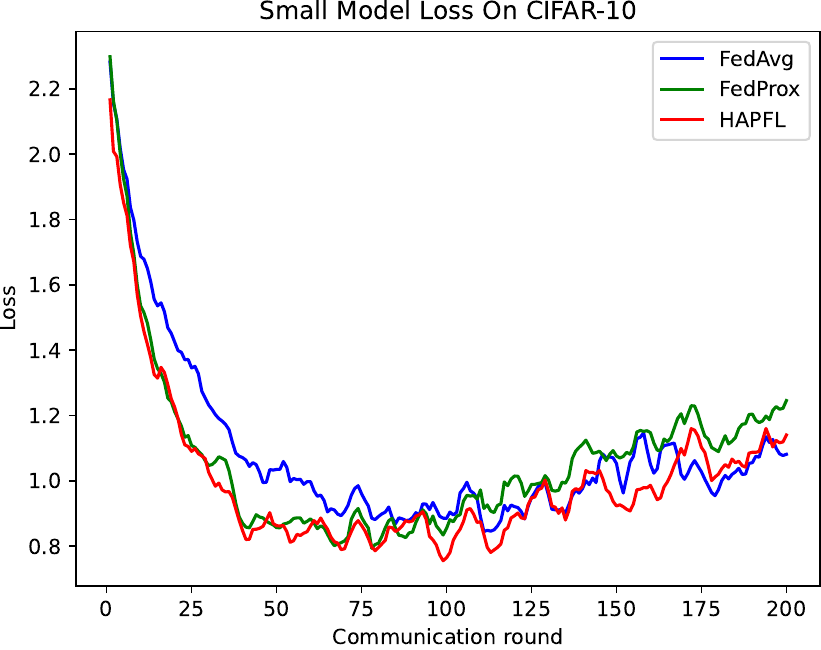}
        \caption{Small model on CIFAR-10}
        \label{fig:loss_image5}
    \end{minipage}
    \vspace{-0.4cm}
\end{figure}

\begin{figure}[htbp]
    \centering
    \begin{minipage}[b]{0.24\textwidth}
        \includegraphics[width=\textwidth]{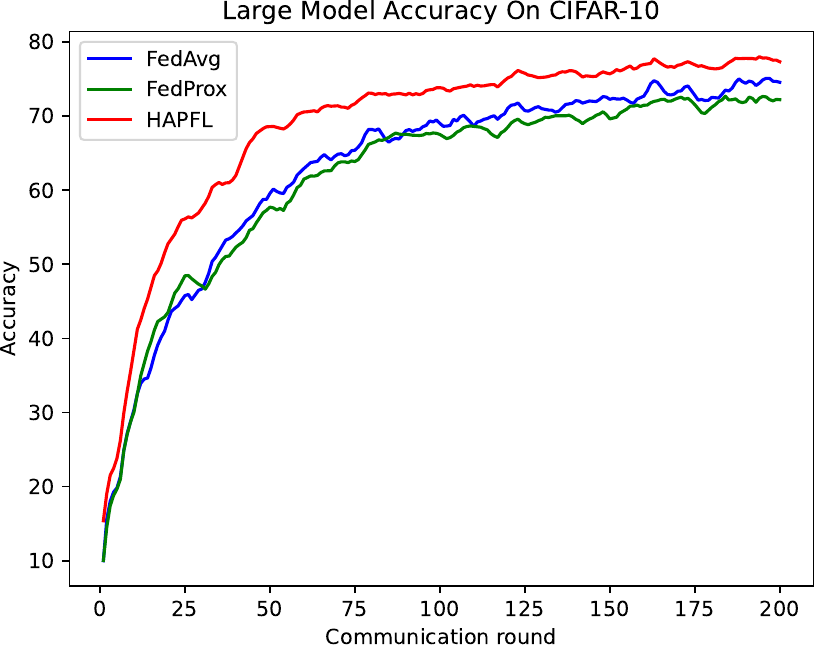}
        \caption{Large model on CIFAR-10}
        \label{fig:image6}
    \end{minipage}
    \hfill 
    \begin{minipage}[b]{0.24\textwidth}
        \includegraphics[width=\textwidth]{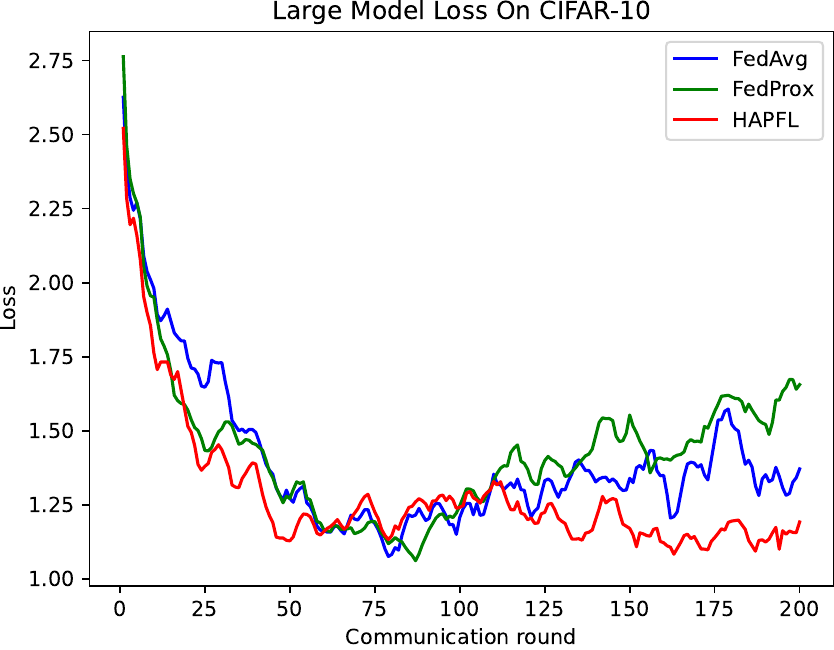}
        \caption{Large model on CIFAR-10}
        \label{fig:loss_image6}
    \end{minipage}
    \vspace{-0.4cm}
\end{figure}

\textbf{CIFAR-10:} Figures \ref{fig:image4}-\ref{fig:loss_image6} present comparative results on the CIFAR-10 dataset. It can be observed from the results that the HAPFL method generally converges faster, with the LiteModel and small model slightly surpassing the baseline in accuracy and the large model exhibiting a notable advantage. This suggests that the mutual learning between the local model and the LiteModel significantly enhances decision-making efficiency. The large and small models significantly outperform the LiteModel in accuracy, with the large and small models achieving approximately 78.0\% accuracy, compared to 74.7\% for the LiteModel. Table \ref{tab:Cifar10 PFL model comparison} further illustrates this trend, showing that HAPFL consistently outperforms pFedMe, with even larger differences in maximum and average accuracy than in Table \ref{tab:MNIST PFL model comparison} for the MNIST dataset.


\begin{figure}[htbp]
    \centering
    \begin{minipage}[b]{0.24\textwidth}
        \includegraphics[width=\textwidth]{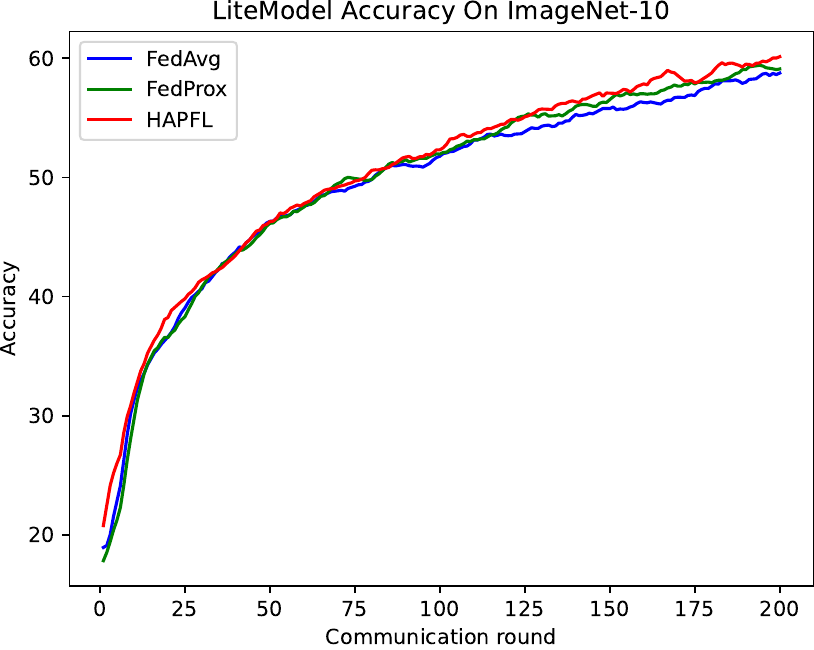}
        \caption{LiteModel on ImageNet-10}
        \label{fig:image7}
    \end{minipage}
    \hfill 
    \begin{minipage}[b]{0.24\textwidth}
        \includegraphics[width=\textwidth]{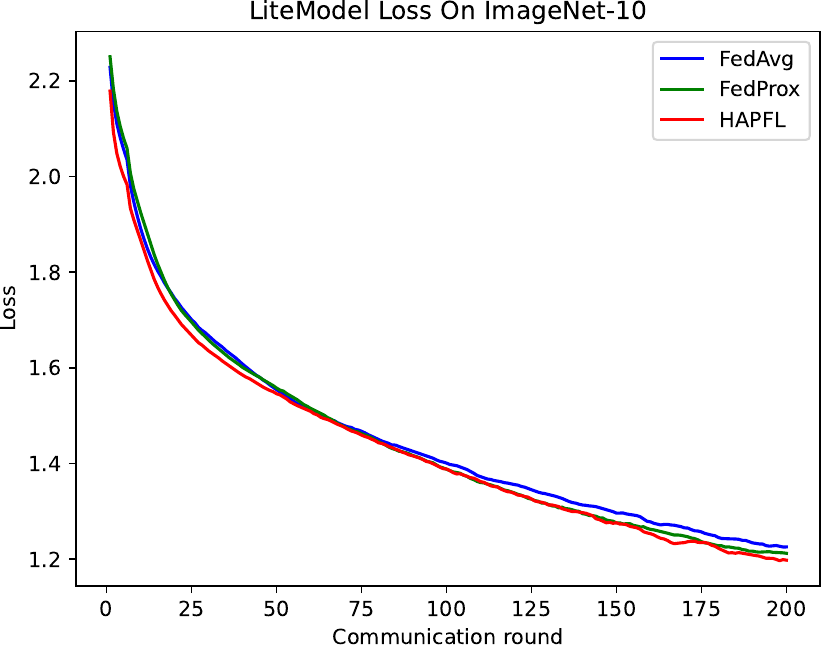}
        \caption{LiteModel on ImageNet-10}
        \label{fig:loss_image7}
    \end{minipage}
    \vspace{-0.4cm}
\end{figure}

\begin{figure}[htbp]
    \centering
    \begin{minipage}[b]{0.24\textwidth}
        \includegraphics[width=\textwidth]{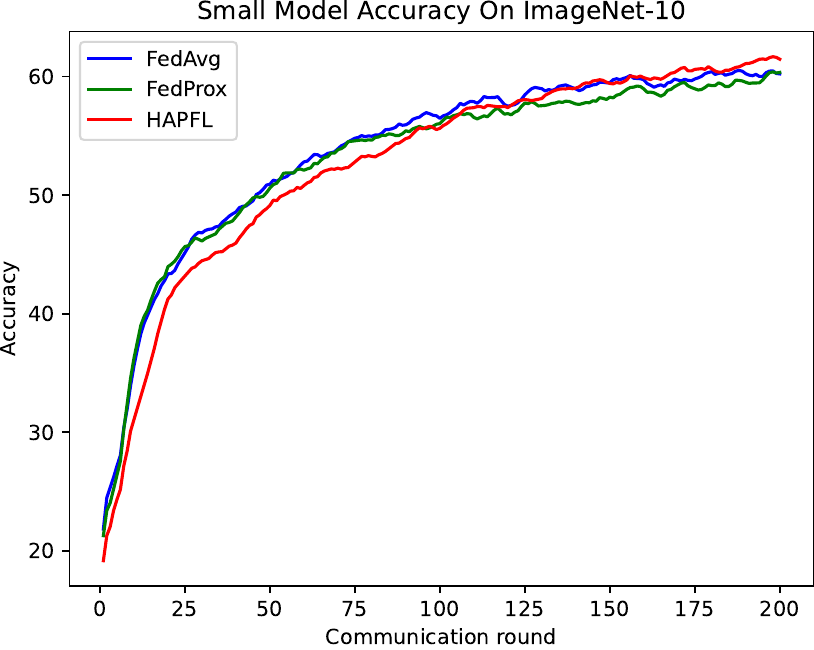}
        \caption{\makebox{Small model on ImageNet-10}}
        \label{fig:image8}
    \end{minipage}
    \begin{minipage}[b]{0.24\textwidth}
        \includegraphics[width=\textwidth]{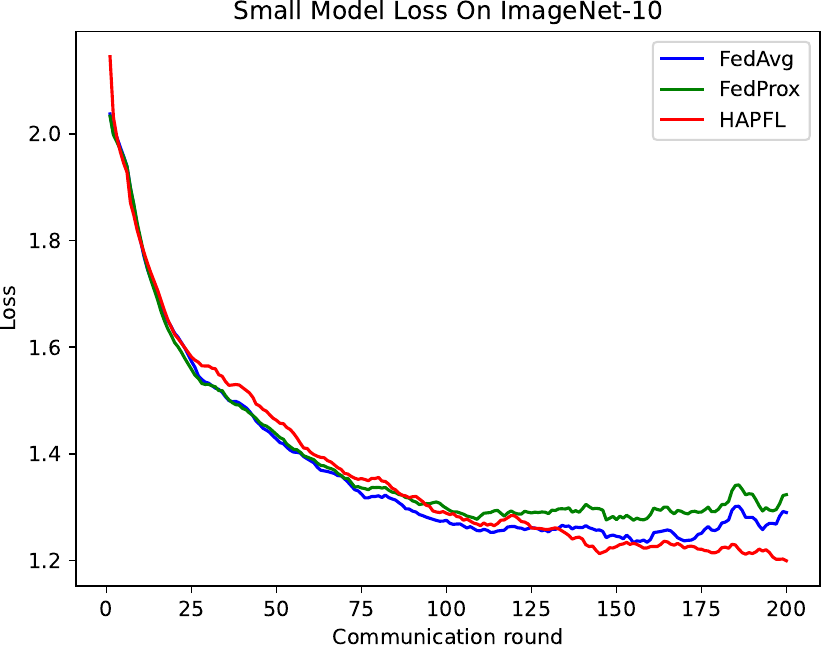}
        \caption{\makebox{Small model on ImageNet-10}}
        \label{fig:loss_image8}
    \end{minipage}
    \vspace{-0.4cm}
\end{figure}

\begin{figure}[htbp]
    \centering
    \begin{minipage}[b]{0.24\textwidth}
        \includegraphics[width=\textwidth]{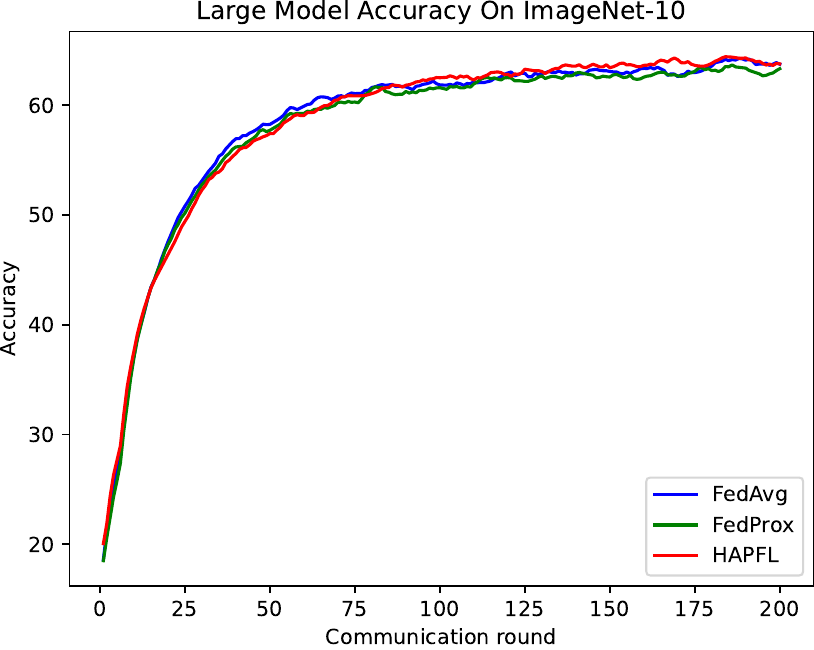}
        \caption{Large model on ImageNet-10}
        \label{fig:image9}
    \end{minipage}
    \hfill 
    \begin{minipage}[b]{0.24\textwidth}
        \includegraphics[width=\textwidth]{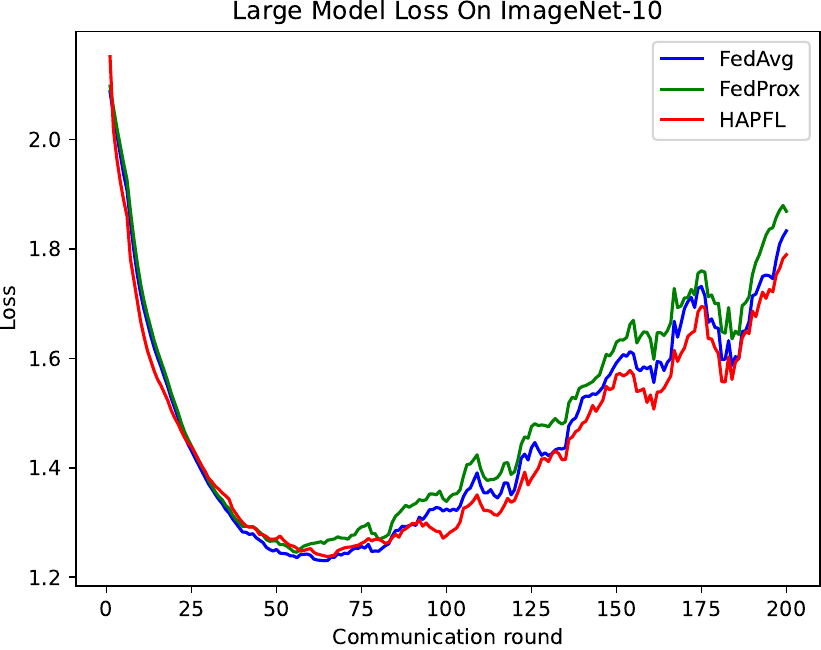}
        \caption{Large model on ImageNet-10}
        \label{fig:loss_image9}
    \end{minipage}
    \vspace{-0.4cm}
\end{figure}

\textbf{ImageNet-10:} Figures \ref{fig:image7}-\ref{fig:loss_image9} show
the comparison results for accuracy and loss on the ImageNet-10 dataset, illustrating the superior performance of HAPFL against the baseline algorithms. Moreover, it can be observed that Figures \ref{fig:image7}, \ref{fig:image8}, and \ref{fig:image9} exhibit a progressive increase in accuracy and convergence speed from the LiteModel to the large model. This underscores the advantage of employing larger models for more complex tasks, as they not only mitigate straggling latency but also exhibit superior performance and faster convergence. 
Specifically, the LiteModel exhibits an upward accuracy trend, with the accuracy reaching approximately 60.2\% after 200 communication rounds.
The small model stabilizes at about 61.0\% accuracy after 150 rounds, while the large model reaches approximately 64.1\% accuracy after 100 rounds.



\begin{table}[htbp]
\centering
\caption{Comparison of personalized model accuracy between HAPFL and pFedMe algorithms on CIFAR-10 dataset}
\label{tab:Cifar10 PFL model comparison}
\resizebox{\linewidth}{!}{
\begin{tabular}{|c|c|c|c|c|c|}
\hline
\multirow{2}{*}{Client} & \multirow{2}{*}{Algorithm} & \multicolumn{2}{c|}{Small Model} & \multicolumn{2}{c|}{Big Model} \\ \cline{3-6} & & Maximum Accuracy & Average Accuracy & Maximum Accuracy & Average Accuracy \\ \hline
\multirow{2}{*}{Client 1} & HAPFL & 72.65\% & 63.90\% & 71.42\% & 57.98\% \\ \cline{2-6} & pFedMe & 24.35\% & 19.36\% & 33.76\% & 28.63\% \\ \hline
\multirow{2}{*}{Client 2} & HAPFL & 74.12\% & 65.81\% & 54.78\% & 54.78\% \\ \cline{2-6} & pFedMe & 54.17\% & 50.76\% & 41.62\% & 41.62\% \\ \hline
\multirow{2}{*}{Client 3} & HAPFL & - & - & 73.69\% & 62.28\%            \\ \cline{2-6} & pFedMe & - & - & 38.7\% & 32.63\% \\ \hline
\multirow{2}{*}{Client 4} & HAPFL & 77.43\% & 68.51\% & - & - \\ \cline{2-6} & pFedMe & 56.6\% & 52.15\% & - & - \\ \hline
\multirow{2}{*}{Client 5} & HAPFL & 76.57\% & 69.38\% & - & - \\ \cline{2-6} & pFedMe & 66.90\% & 62.71\% & - & - \\ \hline
\multirow{2}{*}{Client 6} & HAPFL & - & - & 61.49\% & 50.84\% \\ \cline{2-6} & pFedMe & - & - & 32.06\% & 27.22\% \\ \hline
\multirow{2}{*}{Client 7} & HAPFL & 73.4\% & 65.94\% & 72.82\% & 66.91\% \\ \cline{2-6} & pFedMe & 60.16\% & 54.59\% & 59.98\% & 53.38\% \\ \hline
\multirow{2}{*}{Client 8} & HAPFL & 72.98\% & 72.98\% & 72.0\% & 62.23\% \\ \cline{2-6} & pFedMe & 43.48\% & 43.48\% & 41.13\% & 35.74\% \\ \hline
\multirow{2}{*}{Client 9} & HAPFL & 73.91\% & 64.07\% & - & - \\ \cline{2-6} & pFedMe & 43.98\% & 40.58\% & - & - \\ \hline
\multirow{2}{*}{Client 10} & HAPFL & - & - & 70.12\% & 62.36\% \\ \cline{2-6} & pFedMe & - & - & 54.99\% & 50.85\% \\ \hline
\end{tabular}}
\end{table}

\subsubsection{Straggling Latency and Overall Training Time}
\begin{figure}[htbp]
  \centering
  \begin{minipage}[b]{0.24\textwidth}
    \includegraphics[width=\textwidth]{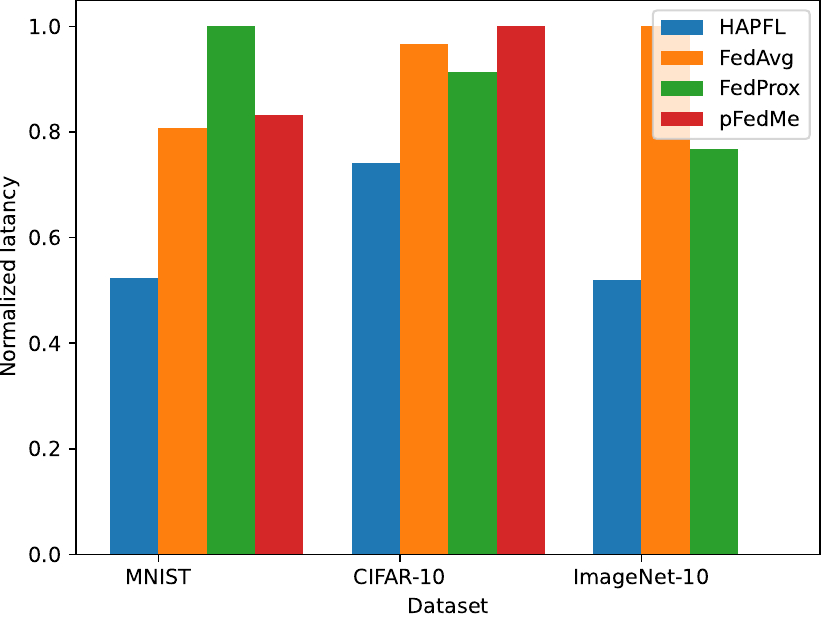}
    \caption{Straggling latency}
    \label{fig:latency}
  \end{minipage}
  \hfill
  \begin{minipage}[b]{0.24\textwidth}
    \includegraphics[width=\textwidth]{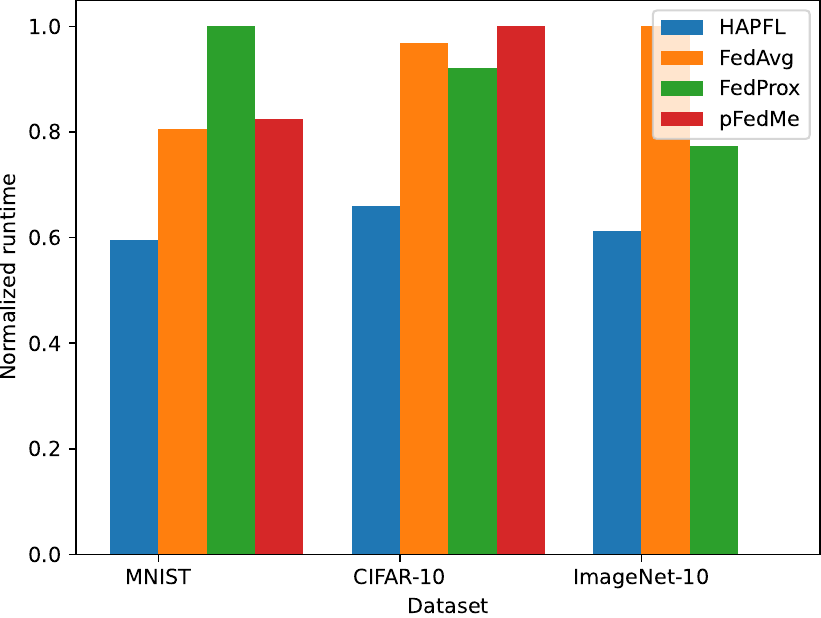}
    \caption{Overall training time}
    \label{fig:runtime}
  \end{minipage}
  \vspace{-0.3cm}
\end{figure}

This subsection assesses the straggling latency (referring to Eq. \ref{training_latency}) and overall training time of the HAPFL method compared to FedAvg, FedProx and pFedMe across three datasets. We present the normalized results for these metrics in Figures \ref{fig:latency} and \ref{fig:runtime}.


Figure \ref{fig:latency} displays the straggling latency comparisons among different methods on various datasets. HAPFL consistently achieves the lowest straggling latency across all datasets. Specifically, on the MNIST dataset, HAPFL reduces straggling latency by 35.2\%, 47.6\% and 37.1\% compared to FedAvg, FedProx and pFedMe, respectively. On the CIFAR-10 dataset, HAPFL shows a reduction in straggling latency of 23.5\%, 19.0\% and 26.0\% compared to FedAvg, FedProx and pFedMe, respectively. On the ImageNet10 dataset, HAPFL achieves a substantial reduction in straggling latency by 48.0\% and 32.2\% compared to FedAvg and FedProx, respectively.

Figure \ref{fig:runtime} illustrates the overall training time of different methods on three datasets. Consistently, HAPFL demonstrates the fastest training speed with the lowest overall training time on each dataset. Specifically, on the MNIST dataset, HAPFL reduces the overall training time by 26.1\%, 40.4\% and 27.7\% compared to FedAvg, FedProx and pFedMe, respectively. For CIFAR-10, the reductions are 31.9\%, 28.4\% and 34.1\%, and for ImageNet-10, HAPFL reduces the overall training time by 38.8\% and 20.9\% compared to the FedAvg and FedProx, respectively.

In addition, we compare the latency of HAPFL with FedDdrl. The results show that HAPFL reduces the training latency by 16.8\% compared to FedDdrl, highlighting the effectiveness of our approach in addressing latency issues.

\subsubsection{Scalability of Heterogeneous Model Allocation}
\begin{figure}[htbp]
  \centering
  \begin{minipage}[b]{0.35\textwidth}
    \includegraphics[width=\textwidth]{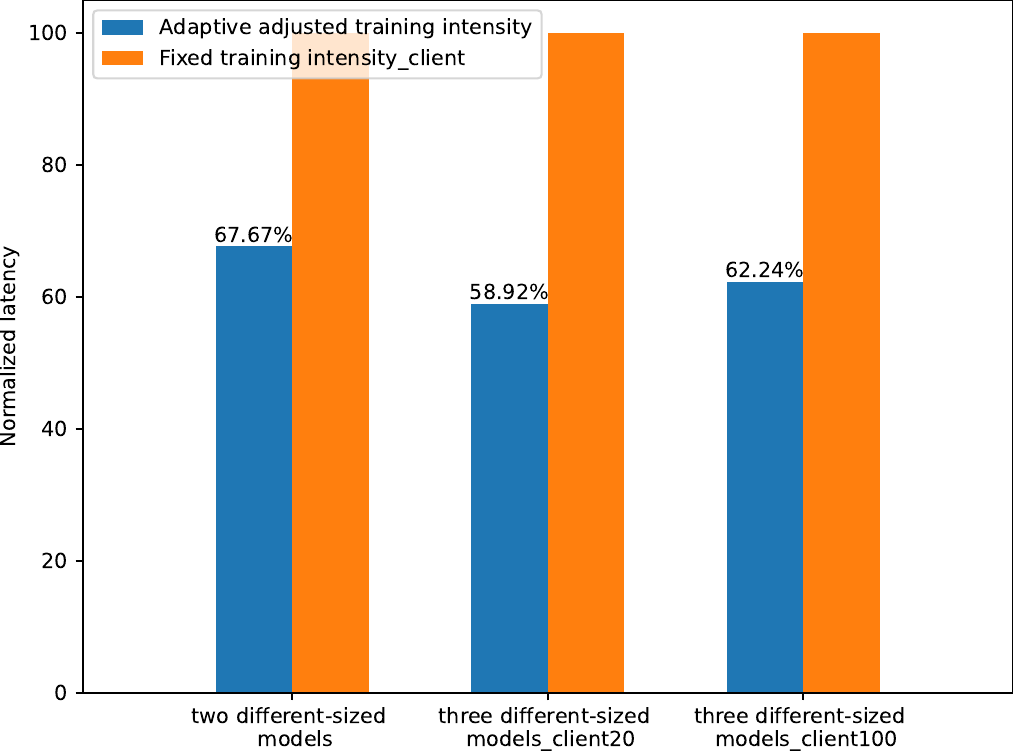}
    \caption{Scalability of model allocation}
    \label{fig:expansionExperiment}
  \end{minipage}
  \vspace{-0.4cm}
\end{figure}

To evaluate the scalability of our RL-based heterogeneous model allocation mechanism, we conducted two sets of extended experiments with different client configurations and performance discrepancies. The first set included 20 clients with performance discrepancies reaching up to 20-fold, while the second set expanded the number of clients to 100, with performance differences escalating to 50-fold. In both cases, we introduced three distinct model sizes: small, medium, and large, presenting increasingly challenging and diverse testing environments. We compared the straggling latency performance of our HAPFL, which utilizes RL-based adaptive adjusted training intensities, against the FedAvg method that employs fixed training intensities.
The results depicted in Figure \ref{fig:expansionExperiment} demonstrate the effectiveness of our model allocation mechanisms in reducing straggling latency. Specifically, the allocation scheme involving two different-sized models resulted in a 32.33\% reduction in straggling latency. When expanded to include three different-sized models in the 20-client setup with 20-fold performance discrepancies, this strategy further improved performance, reducing the average straggling latency by 41.08\%. In the 100-client setup with 50-fold performance discrepancies, the three-model allocation achieved a 37.76\% reduction in training latency. Although this is slightly lower than the 20-client setup, it outperforms the two-model, 10-client configuration. This result suggests that the availability of three distinct model sizes allows for finer-grained allocation, while larger client discrepancies demand more precise distribution, which naturally limits the reduction in straggling latency.
These results not only prove the adaptability and effectiveness of our model allocation scheme in more complex and varied FL environments but also highlight its significant scalability, enabling it to efficiently operate within larger-sized FL systems.

\subsubsection{Ablation Study}
\begin{figure}[htbp]
  \centering
  \begin{minipage}[b]{0.35\textwidth}
    \includegraphics[width=\textwidth]{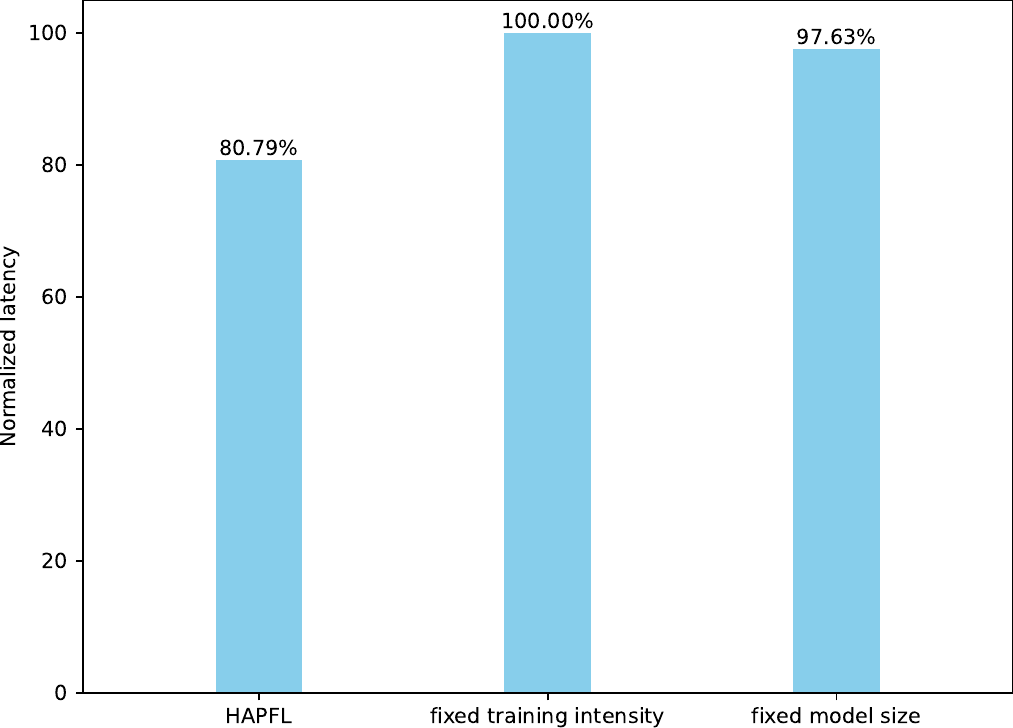}
    \caption{Ablation studies}
    \label{fig:ablationStudies}
  \end{minipage}
  \vspace{-0.4cm}
\end{figure}

To verify the contributions of each component in the HAPFL method, we conducted ablation studies with two separate experiments. In one experiment, clients were assigned models of fixed size without varying the training intensity. In the other, clients were assigned fixed training intensity without adjusting model sizes. The results are shown in Figure \ref{fig:ablationStudies}. Compared to using fixed training intensity, HAPFL achieved a 19.21\% reduction in training latency. Similarly, compared to using fixed model size, HAPFL reduced training latency by 17.23\%. While the introduction of additional reinforcement learning models incurs some overhead, the improvements in performance provided by our method far outweigh these costs. By combining both strategies, HAPFL effectively mitigated client performance disparities, leading to a significant decrease in overall training latency. These findings indicate that both assigning different model sizes and adjusting training intensity are essential for enhancing the performance of HAPFL.

\section{Conclusion}
\label{conclusion}
This paper proposes the HAPFL method, designed specifically to mitigate the ``straggler problem" inherent in synchronous FL environments characterized by heterogeneous client capacities. Our approach dynamically allocates diverse models and training intensities to clients based on their performance, effectively minimizing straggling latency. 
In the HAPFL method, each participating client operates with a uniformed LiteModel alongside a customized heterogeneous local model. These models engage in mutual learning through knowledge distillation, enhancing their performance. 
Subsequent aggregation of all models is conducted using a weighted method that considers both the information entropy of client data and model accuracy, ensuring optimal integration of learned features.
Comprehensive experiments on three public datasets, MNIST, CIFAR-10, and ImageNet-10, demonstrate that, compared with baseline algorithms, HAPFL not only significantly improves model accuracy but also optimizes straggling latency and overall training time. 


\bibliographystyle{unsrt}
\bibliography{sample}

\end{document}